\documentclass{article}

\usepackage{arxiv}

\usepackage{mathtools}    
\usepackage{amssymb}
\usepackage{bm}

\usepackage{times}
\usepackage{textcomp}
\usepackage{soul}
\usepackage{graphicx}
\usepackage{svg}
\usepackage{wrapfig}
\usepackage{float}
\usepackage{epstopdf}     
\usepackage{xcolor}

\usepackage{tikz}
\usepackage{adjustbox}

\usetikzlibrary{shapes, arrows.meta, positioning, calc, shadows, backgrounds, fit}
\usepackage{smartdiagram}

\usepackage{booktabs}
\usepackage{tabularx}
\usepackage{multirow}
\usepackage{makecell}
\usepackage{colortbl} 
\usepackage{hhline}
\usepackage{siunitx,ragged2e}
\usepackage{tcolorbox}
\usepackage[table,xcdraw]{xcolor}
\usepackage{dashrule}

\usepackage{caption}
\usepackage{subcaption}

\usepackage{algorithm}
\usepackage{algpseudocode}

\usepackage{enumitem}
\usepackage{enumerate}

\usepackage{pdfrender}

\usepackage{pifont}
\usepackage{bbding}
\usepackage{comment}
\usepackage{fancyvrb}
\usepackage{blindtext}
\usepackage{setspace}
\usepackage{url}
\usepackage[english]{babel}

\usepackage{hyperref}
\hypersetup{colorlinks}
\usepackage{cleveref}
\usepackage{wrapfig}

\usepackage{natbib}
\setcitestyle{authoryear,open={(},close={)}}

\definecolor{llmfree}{RGB}{180,180,180}
\definecolor{llmdriven}{RGB}{120,170,255}
\definecolor{retrieval}{RGB}{70,130,200}

\definecolor{darkpastelgreen}{rgb}{0.06, 0.70, 0.24}
\definecolor{electriccrimson}{rgb}{1.0, 0.0, 0.25}
\definecolor{navyblue}{rgb}{0.0, 0.0, 0.75}
\newcommand{\ie}{\textit{i.e.}}
\newcommand{\eg}{\textit{e.g.}}

\newcommand{\algacro}{\textsc{AppSelectBench}{}}

\title{\algacro{}: Application-Level Tool Selection Benchmark}

\author{
Please refer to the full author list in Appendix~\ref{appendix:authorlist}.\\
{Microsoft}
}

\begin{document}

\maketitle

\begin{abstract}
Computer Using Agents (CUAs) are increasingly equipped with external tools, enabling them to perform complex and realistic tasks. 
For CUAs to operate effectively, \textit{application selection}, which refers to deciding which application to use before invoking fine-grained tools such as APIs, is a fundamental capability. 
It determines whether the agent initializes the correct environment, avoids orchestration confusion, and efficiently focuses on relevant context. 
However, existing benchmarks primarily assess fine-grained API selection, offering limited insight into whether models can reason across and choose between different applications. 
To fill this gap, we introduce \algacro{}, a comprehensive benchmark for evaluating {application selection} in CUAs.  
\algacro{} contains a novel user task generation pipeline that produces realistic, diverse, and semantically grounded user intents at scale, together with unified evaluation protocols covering random, heuristic, zero-shot, few-shot, and retrieval-augmented-settings. 
\algacro{} covers one hundred widely used desktop applications and includes more than one hundred thousand realistic, diverse, and semantically grounded user tasks.
Extensive experiments across both closed-source and open-source large language models reveal systematic strengths and weaknesses in inter-application reasoning, showing that even the most capable models still struggle to make consistent application choices. 
Together, these results establish \algacro{} as a foundation for studying and advancing application level reasoning, an essential yet underexplored capability of intelligent CUAs. The source is available at \url{https://microsoft.github.io/appselectbench/}.
\end{abstract}

\section{Introduction}

Computer Using Agents (CUAs) have demonstrated remarkable proficiency in tool-augmented reasoning, where agents invoke external  APIs, plugins, or operating system utilities to accomplish complex goals~\citep{schick2023toolformerlanguagemodelsteach,ac1f09077393404a8bea5141d8710259,wang2024mintevaluatingllmsmultiturn,Braunschweiler2025ToolReAGt}, accompanying with measuring their tool selection ability becomes important. Recent benchmarks such as API-Bank~\citep{li2023apibankcomprehensivebenchmarktoolaugmented,patil2024gorilla} and ToolBench~\citep{xu2023tool,qin2023toolllm} have therefore centered on evaluating models’ ability to select and execute API calls correctly within pre-defined toolsets. In contrast, real human and computer interaction rarely begins at the API level. 
Users typically reason at the \textit{application level}, deciding whether to use a spreadsheet, a web browser, or an image editor, before invoking specific operations within that application. 
Such \textit{application selection} governs the human orchestration of computer use and forms a crucial bridge between user intent and  fine-grained tool execution. 

Despite the central role of application selection in human-computer interaction workflows, it has received limited attentions in existing agentic tool-use research. This gap naturally raises a key question.

\textit{\textbf{Is it important to establish and measure the application-selection capability for CUAs?}}

From our perspective, the answer is \texttt{YES} due to the following reasons.

\textbf{Provide good initialization of the environment.} 
Selecting the appropriate application provides a good initialization for CUA, favoring it a suitable environment for accomplishing the task. Such initialization often leads to substantial improvements in downstream execution accuracy and task completion efficiency~\citep{}.

In CUA benchmarks such as OSworld ~\citep{xie2024osworld} and WAA~\citep{bonatti2024windows}, applications related to a task are often preloaded to reduce the ambitiuity of the task. However, in real world scenario, being able to perform a task from end to end, especially starting from selecting the right application is highly valuable.    

\textbf{Reduce confusion of the orchestrator.} 
Without clear application-level grounding, an orchestrator exposed to a large number of fine-grained API calls can easily become confused and deliver incorrect or inconsistent decisions. 
Application-level reasoning provides structural clarity that stabilizes orchestration and improves performance.

\textbf{Enhance context efficiency.} 
Resolving the application selection first allows CUAs to focus on relevant APIs and operations within the chosen environment while ignoring irrelevant tools, improving both reasoning efficiency.

Building such an application-level tool selection benchmark is \textit{non-trivial}. Unlike API-level evaluation, it demands \textit{(i)} a large and diverse coverage of applications with overlapping functionalities, \textit{(ii)} diverse and realistic user-task instructions that reflect authentic intents rather than synthetic command templates, and \textit{(iii)} structured representations that can capture temporal or logical dependencies across applications. Moreover, ground truth application might be ambiguous since many tasks admit multiple valid applications. All these make both dataset construction and evaluation design technically challenging.

\begin{figure}[t]
    \centering
    \includegraphics[width=0.99\linewidth]{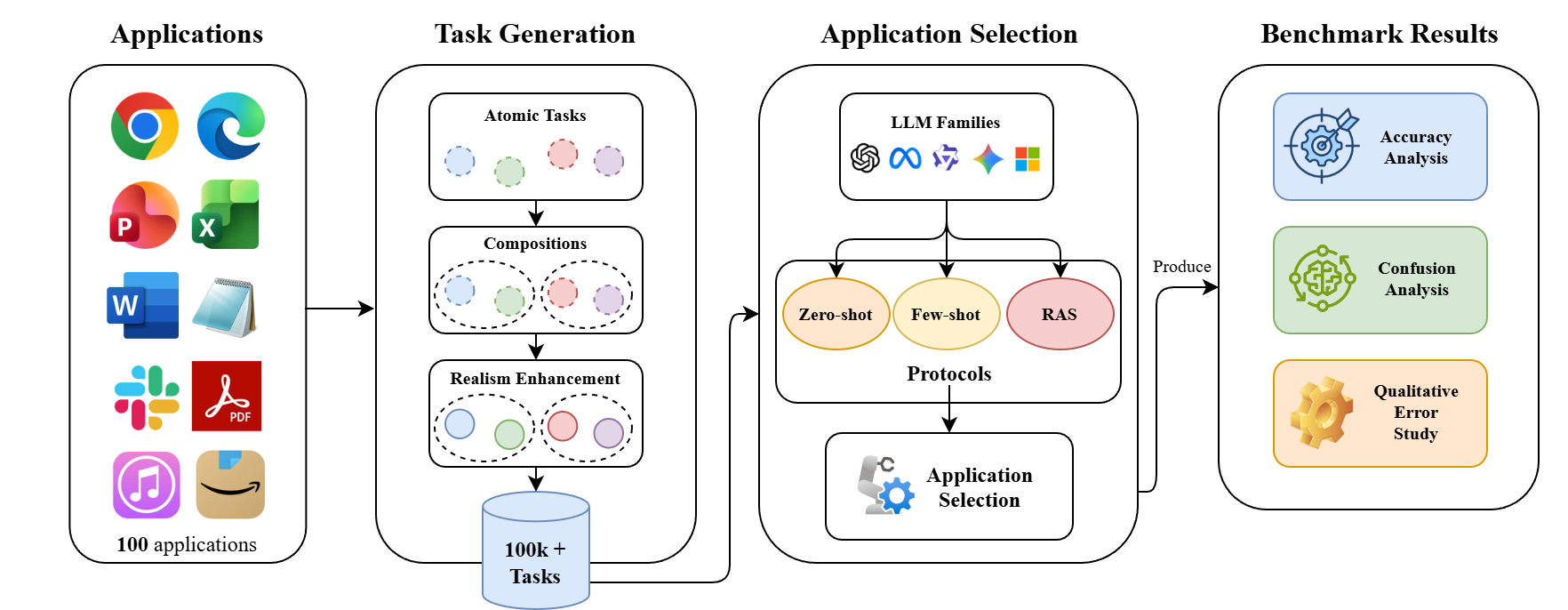}
    \caption{The overview of the \algacro{} framework, including three core stages: (1) a multi-step generation pipeline producing over 100k realistic tasks; (2) a parallelized tool-selection core where multiple LLMs are tested against various protocols; and (3) the resulting in-depth performance analysis.}
    \label{fig:overview}
\end{figure}

To address these challenges, we introduce \algacro{}, the first comprehensive benchmark explicitly targeting \textit{application-selection} in computer-using agents to the best of our knowledge. \algacro{} comprises two main components: 
\textit{(i)} a user-task generation pipeline that synthesizes diverse, realistic, and semantically grounded user tasks covering a broad range of applications, and 
\textit{(ii)} unified evaluation protocols that test various model families and prompting strategies to systematically measure their application-selection ability. 

Our main contributions are fourfold:

\begin{itemize}[leftmargin=12pt]
    \item \textbf{Benchmark Design.} We introduce \algacro{}, to the best of our knowledge, the first benchmark dedicated to evaluating CUAs’ capability in selecting appropriate applications for open-ended user intents, advancing the understanding of how agents translate natural-language goals into executable application choices.
    
    \item \textbf{User Task Generation Pipeline.} We propose a novel multi-stage pipeline that composes atomic tasks into realistic user tasks and narrates them into natural language. This pipeline reliably produces diverse, high-quality, and verifiable application–task pairs that can serve multiple evaluation and training purposes.
    
    \item \textbf{Unified Evaluation Protocols.} We establish standardized application-selection regimes, including random, heuristic, zero-shot, few-shot, and retrieval-augmented settings. These strategies provide comprehensive ways to assess both closed-source and open-source models under consistent conditions.
    
    \item \textbf{Comprehensive Analysis.} We conduct extensive evaluations across diverse representative LLMs, accompanied by detailed confusion and error analyses that uncover systematic strengths and weaknesses in inter-application reasoning. The results reveal that existing models still exhibit a substantial performance gap, underscoring the significant challenges that remain for advancing CUA research.
\end{itemize}

Together, these advances establish \algacro{} as a sound foundation for studying application-level reasoning, bridging the gap between user-intent understanding and concrete tool selection in computer-using agents.

\section{Related Work} 

\paragraph{API-Level Tool Selection.} 
Recent research has demonstrated the ability of LLMs to use external tools to improve their capability and reliability~\citep{ qu2025tool, huang2023metatool}. These tools can augment the LLMs in categories such as search~\citep{schick2023toolformer}, math~\citep{bulusu2024mathviz}, or coding~\citep{gao2023pal}. To accurately evaluate and compare these new capabilities, a number of tool selection benchmarks have been developed~\citep{guo2024stabletoolbench, li2023api, zhuang2023toolqa, patil2024gorilla, tang2023toolalpaca}.
These existing tool use paradigms predominantly evaluate API-level competence, \ie, whether a model can correctly invoke functions within a pre-selected application (\eg, \texttt{SUM(A1:A10)} in a spreadsheet). In contrast, our work addresses a higher-level reasoning problem: deciding which application to use before deciding how to use it. This distinction parallels human workflow planning, where one first determines whether to open a spreadsheet, document editor, or browser then performing fine-grained operations within it. The conceptual differences between API-level and application-level selection are summarized in Table~\ref{tab:api-vs-app-level}.

\begin{table}[h]
\centering
\small
\caption{Comparison of API-level vs. application-level tool selection.}
\label{tab:api-vs-app-level}
\begin{tabular}{lll}
\toprule
\textbf{Dimension} & \textbf{API-Level Selection} & \textbf{Application-Level Selection} \\
\midrule
Scope & Intra-tool function calls & Inter-application choice \\
Input & User Task + Pre-selected App & User Task Only \\
Output & Sequence of API calls & Set of suitable applications \\
Core Challenge & Argument binding, sequencing & High-level intent matching \\
Example & \texttt{SUM(range)} in Excel & Excel vs. Calculator \\
\bottomrule
\end{tabular}
\end{table}

\paragraph{Computer Use Agents (CUAs).}
Computer-using agents (CUAs) extend tool-augmented LLMs into full operating environments capable of GUI interaction through actions like clicking and typing~\citep{wang2025opencua, wang2025ui,zhang2025phi}, with recent work integrating high-level tool calls to improve efficiency and reliability~\citep{yang2025ultracua}. 
Nevertheless, existing CUA benchmarks~\citep{xie2024osworld,bonatti2024windows,hui2025winspot}, typically evaluate end-to-end task completion within preloaded environments, where the relevant application is already opened and ready to use. This design simplifies execution but bypasses a fundamental step in realistic workflows: deciding which application to launch to accomplish a user goal.
In practice, agents often fail when the starting environment is unset or when multiple plausible applications could be used for the same task, underscoring the need to explicitly assess application selection ability. Our benchmark targets precisely this missing layer, providing a controlled yet realistic setup for evaluating inter-application reasoning independently of execution fidelity.

\begin{figure}[h]
	\centering
	\includegraphics[width=0.98\linewidth]{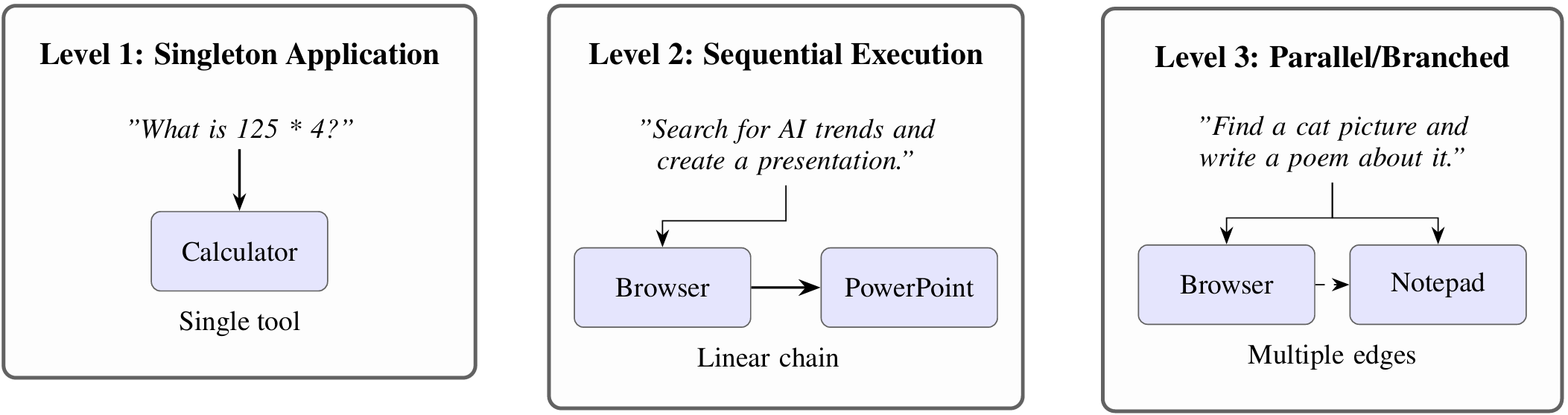}
	\caption{The three levels of task complexity: (1) \textbf{Singleton Application} for tasks solvable by a single tool; (2) \textbf{Sequential Execution} represented by a linear chain (\eg, Browser $\rightarrow$ PowerPoint); (3) \textbf{Parallel or Branched Execution} with multiple outgoing or incoming edges.}
	\label{fig:task_levels}
\end{figure}

\section{Problem Definition}

We formalize the application-level tool selection problem as a \textbf{graph-structured prediction} task, where an agent must map a natural-language user intent to a directed graph of appropriate applications and their dependencies.

Given a user task description $\mathcal{U}$ expressed in natural language and a candidate application set $\mathcal{T} = \{ t_1, t_2, \ldots, t_n \}$, the goal is to construct a directed graph $G({\mathcal{U}}) = (V, E)$ that encodes the structure of tool usage required to accomplish $\mathcal{U}$. Formally, the task is to learn a function
\begin{equation}
f: \mathcal{U} \rightarrow \mathcal{G}(\mathcal{T}),
\end{equation}
where $\mathcal{G}(\mathcal{T})$ denotes the space of all directed graphs composed of vertices in $\mathcal{T}$ and edges $E \subseteq \mathcal{T} \times \mathcal{T}$, thereby the predicted tool structured graph $G$ belonging to $\mathcal{G}$, \ie, $G({\mathcal{U}})\in \mathcal{G}$. Each vertex $v_i \in V$ represents an instantiated application, while each directed edge $e_{ij} = (v_i, v_j)$ represents a temporal, logical, or data dependency between two tool invocations. This formulation naturally captures a wide range of tool interaction patterns:
\begin{itemize}
    \item \textbf{Singleton application}, representing tasks solvable by a single tool;
    \item \textbf{Sequential execution}, represented by linear chains (\eg, Browser $\rightarrow$ PowerPoint);
    \item \textbf{Parallel or branched execution}, represented by multiple outgoing or incoming edges.
\end{itemize}

The graph-based perspective generalizes both unordered and ordered multi-label formulations, and models concurrent, and interdependent tool usage for realistic Computer-Using-Agent (CUA) planning and reasoning.

\paragraph{Focus and Scope.} In this work, we focus on the \textbf{singleton application selection setting}, where each task is accomplished by a single application or a few alternatives. Extensions to multi-application scenarios involving sequential or parallel tool usage are left for future versions of this study.

\begin{figure}[H]
    \centering
    \includegraphics[width=0.5\linewidth]{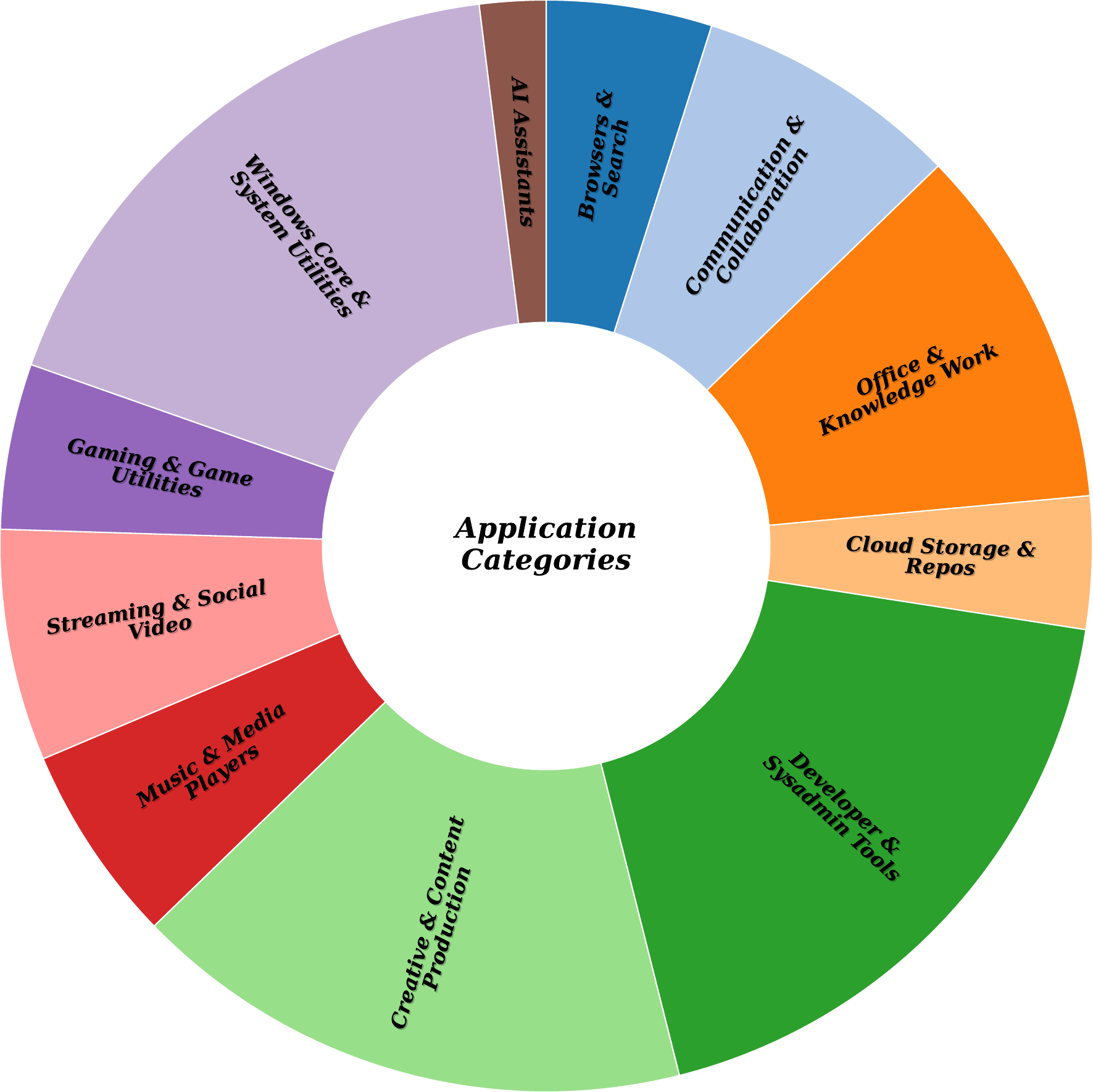}
    \caption{Distribution of the 100 desktop applications in AppSelectBench across 12 high-level categories.}
    \label{fig:application_coverage}
\end{figure}

\section{Benchmark Design}

Our benchmark targets the \emph{application-level} decision problem: given a natural-language task, select the most appropriate (or a set of valid alternatives) desktop application family, and when appropriate, in a structured manner. The design emphasizes the coverage of applications to the computer-use and the representative user tasks with realistic, associated with the verifiable ground truth.

\subsection{Application Coverage and Category}\label{sec:application_coverage_and_domains}

To ensure fairness and representativeness in evaluation, we systematically curate a diverse set of applications that span the spectrum of everyday computer use. Applications are grouped into twelve high-level categories that reflect distinct modes of human–computer interaction, as summarized in \autoref{fig:application_coverage} detailing in Appendix~\ref{sec:application_domain}. These categories span \textbf{Browsers and Search} (web access and information retrieval), \textbf{Communication and Collaboration} (chat, meetings, messaging, remote access), \textbf{Office and Knowledge Work} (documents, spreadsheets, presentations, notes, task management, and automation), \textbf{Cloud Storage \& Repos} (file synchronization and code hosting), \textbf{Developer and Sysadmin Tools} (IDEs, shells, containers, networking, and provisioning), \textbf{Creative and Content Production} (image, audio, and video creation and editing), \textbf{Music and Media Players} (audio and video playback), \textbf{Streaming and Social Video} (consumer video platforms), \textbf{Gaming and Game Utilities} (game launchers, overlays, and assistive tools), \textbf{Windows Core Apps and System Utilities} (file management, configuration, diagnostics, and troubleshooting), and \textbf{AI Assistants} (assistant-style copilots).

Each application group contains a diverse mixture of software ecosystems to capture both overlapping and distinct functionalities. For instance, \emph{Word}, \emph{Notion}, and \emph{Overleaf} all support document authoring but differ in interaction modality and integration depth, while \emph{Photoshop} and \emph{GIMP} represent parallel paradigms for creative image editing. This diversity introduces natural ambiguity, enabling the benchmark to evaluate whether models can reason about \emph{functional equivalence}, \emph{contextual appropriateness}, and \emph{cross-application substitution}, rather than relying on one-to-one lexical associations between task descriptions and tools.

\subsection{Evaluation Metric}\label{sec:evaluation_metrics}

To provide a comprehensive assessment of decision quality, we employ two complementary metrics, accuracy and confusion analysis, each capturing distinct aspects of performance.

\paragraph{Tool selection accuracy.} Accuracy measures the proportion of user tasks for which the predicted application belongs to the annotated set of valid solutions. Because many tasks can be completed using multiple applications, a prediction is considered correct if it matches any of these valid options rather than a single canonical one.

\paragraph{Confusion analysis.} To understand systematic errors, we construct category-level confusion matrices that reveal patterns of misclassification. These analyses expose, for example, when models repeatedly confused productivity tools with overlapping functionality or misidentify communication platforms across families. Such insights enable finer-grained interpretation of model behavior beyond accuracy, clarifying where application boundaries remain ambiguous.

\subsection{Tool Selection Protocol}\label{sec:tool_selection_protocol}

We evaluate baseline systems and large language models (LLMs) under a set of unified protocols to measure their ability for {application-level tool selection}. The setup covers four regimes, \ie, LLM-free \emph{random and heuristic rule-based baselines}, LLM-driven \emph{zero-shot prompting}, \emph{few-shot prompting}, and \emph{retrieval-augmented selection}, to progressively incorporate prior knowledge from base models and external context.

To establish the baseline of application selection, we introduce two LLM-free baselines. 
\paragraph{Random Selector.} Select an application uniformly at random from the available candidate set. This schema is expected to provide a lower-bound reflecting chance-level performance.

\paragraph{Rule-based Heuristic.} Employ semantic matching between task descriptions and predefined application-specific lexicons (e.g., \textit{sum}, \textit{cell}, \textit{spreadsheet} $\rightarrow$ Excel; \textit{slide}, \textit{presentation} $\rightarrow$ PowerPoint). While simplistic, this baseline approximates deterministic retrieval systems commonly found in traditional virtual-assistant pipelines.

We then explore three LLM-driven prompting strategies with increasing levels of supervision and contextual grounding. Unlike the LLM-free baselines, these strategies allow LLMs to leverage their built-in knowledge about applications and task semantics. By gradually adding guidance—from no examples (zero-shot), to a few demonstrations (few-shot), to retrieved similar cases (retrieval-augmented), we examine how different sources of prior information influence the LLMs' ability to select the right application for a given task.

\paragraph{Zero-Shot Prompting.} 
In the zero-shot regime, the model directly receives the user task description and is asked to predict the most appropriate application(s) without any exemplars. This setting measures the model’s inherent world knowledge and generalization ability regarding desktop applications and their functionalities. 
\begin{tcolorbox}[
  colback=blue!10!white,    
  colframe=blue!60!black,   
  title=\textbf{Zero-Shot Prompting Example},
  boxrule=0.4pt,           
  arc=2pt,                 
  left=4pt, right=4pt, top=2pt, bottom=2pt
]
User Task: \textcolor{orange}{Calculate the total sales by region.} \\
Which application would you use?
\end{tcolorbox}
An example of zero-shot prompting is present as the above, where the LLMs are only aware of the \textcolor{orange}{user task description}.

\paragraph{Few-Shot Prompting.}
To provide task–application correspondences, we augment the input prompt with a small number of annotated examples presenting typical mappings, \eg, ``\texttt{calculate the sum of a column $\to$ Excel}" and ``\texttt{create a slide deck for a talk $\to$ PowerPoint}". This paradigm evaluates whether in-context clues can guide models to recognize functional analogies and improve selection accuracy.

\begin{tcolorbox}[
  colback=blue!10!white,    
  colframe=blue!60!black,   
  title=\textbf{Few-Shot Prompting Example},
  boxrule=0.4pt,           
  arc=2pt,                 
  left=4pt, right=4pt, top=2pt, bottom=2pt
]
Reference Example: \textcolor{darkpastelgreen}{Create a bar chart from a table → Excel.}\\
Reference Example: \textcolor{darkpastelgreen}{Compose an email with attachments → Outlook.}\\
Reference Example: \textcolor{darkpastelgreen}{Trim a short video clip → ClipChamp.}\\
User Task: \textcolor{orange}{Add captions to an image.}\\
Which application would you use?
\end{tcolorbox}
Few-shot results illustrate how effectively LLMs can generalize from a small number of \textcolor{darkpastelgreen}{reference examples}, improving their ability to associate natural-language \textcolor{orange}{user task description} with the most relevant applications.

\paragraph{Retrieval-Augmented Selection (RAS).}
In this setting, the model is further provided with structured descriptions of each application, including its core capabilities, functionalities, and typical usage scenarios, retrieved from an external knowledge base. This augmentation not only enriches the model’s understanding of available tools but also broadens the candidate space by offering more comprehensive application information. The retrieved context serves as lightweight factual grounding, allowing us to test whether such augmentation improves decision reliability and robustness under varying task descriptions.

\begin{tcolorbox}[
  colback=blue!10!white,    
  colframe=blue!60!black,   
  title=\textbf{Retrieval-Augmented Selection (RAS) Example},
  boxrule=0.4pt,           
  arc=2pt,                 
  left=4pt, right=4pt, top=2pt, bottom=2pt
]
Application tool list:\\
\textcolor{navyblue}{Excel: Supports tabular editing, formulas, data visualization.}\\
\textcolor{navyblue}{Word: Supports text editing, document formatting.}\\
\textcolor{navyblue}{...}\\
\textcolor{navyblue}{Chrome: Web browsing, video search.}\\
User Task: \textcolor{orange}{I want to know the stock price of Microsoft.}\\
Which application would you use?
\end{tcolorbox}

As illustrated above, LLMs are given access to a tool list enriched with high-level descriptions of each application. This augmented context can assist models in making more informed decisions about which tool to use for a given task. At the same time, it also tests the model’s reasoning ability, specifically, whether it can correctly interpret and utilize the provided external knowledge to guide its selection.

\section{User Task Generation}

\begin{figure}[ht]
    \centering
    \includegraphics[width=0.9\linewidth]{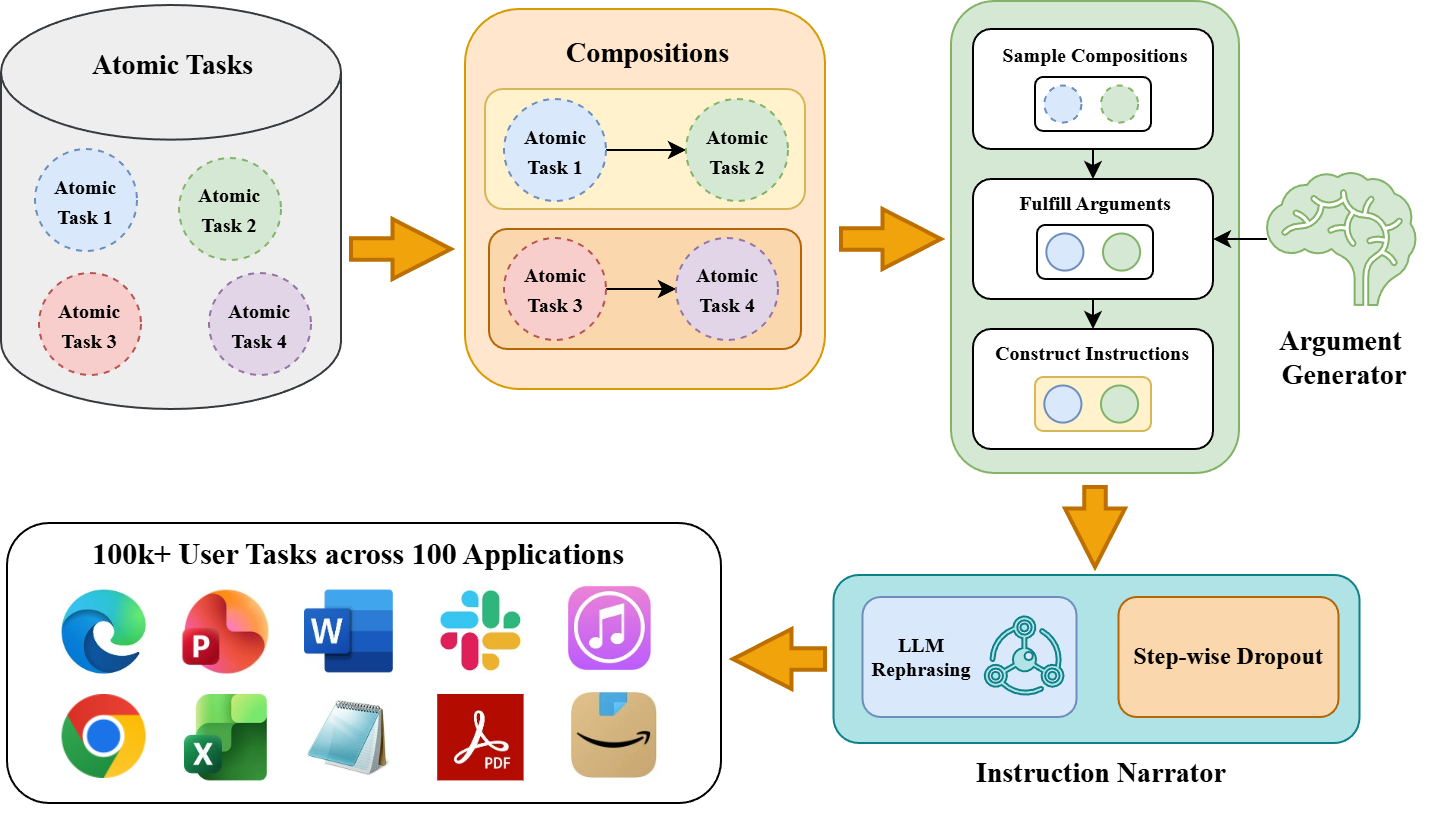}
    \caption{User task generation pipeline that composes atomic actions into workflows and narrates them into natural-language instructions.}
    \label{fig:user_task_generation_overview}
\end{figure}

In our user-task generation pipeline (see \autoref{fig:user_task_generation_overview}), we proceed in four stages. First, we curate an application-specific atomic task database (\eg, \texttt{SUM(column)} for spreadsheets) with explicit argument schemas. Second, a composition engine samples and composes these primitives into higher-level workflows (\eg, \texttt{OpenPowerPoint} $\rightarrow$ \texttt{CreatePresentationFromTemplate}). Third, an argument generator instantiates realistic values to concretize the abstract workflows. Finally, an instruction narrator synthesizes natural-language user-task instructions by integrating step-wise and atomic-task descriptions, yielding fluent and realistic prompts.

\paragraph{Atomic Task} is defined as the minimal unit of user intent or interaction that a human performs when operating a computer system toward a specific purpose. Formally, it represents the smallest semantically meaningful operation that cannot be further decomposed without losing its functional integrity. Atomic tasks generalize across applications and environments. Some examples are like \texttt{CreateNewPresentation}, \texttt{OpenFile}, \texttt{TypeText}, or \texttt{CalculatorSwitchMode}. The curation requires the dedicated efforts of the mixture of GPT-driven curation and human annotations. As a result, we establish a database that consists of three thousands atomic tasks across 100 applications shown in  Section~\ref{sec:application_coverage_and_domains}.

\paragraph{Atomic Task Composition} models how multiple atomic tasks are organized to achieve higher-level user goals. It captures both intra-application semantics and inter-application semantics. Intra-application semantics describe how atomic tasks within the same application interact to form coherent workflows (\eg, \texttt{NodePadOpenFile} $\rightarrow$ \texttt{NodePadEditContent} $\rightarrow$ \texttt{NodePadSaveFile} in Notepad). Inter-application semantics capture how tasks span across different applications to accomplish composite goals (\eg., \texttt{ExcelExportChart} $\rightarrow$ \texttt{PowerPointInsertTable}).  The composition engine samples and connects atomic tasks under logical, temporal, and functional constraints to ensure that generated compositions are semantically valid, contextually coherent, and reflective of realistic user behavior. 

\paragraph{Argument Generation} grounds abstract atomic tasks with concrete, context-aware parameters to specify  them. Some atomic tasks are associated with one or more arguments that define their operational details, such as object names, file paths, numerical values, or textual content. For example, the atomic task \texttt{PowerPointSaveFileAs} may take an argument specifying the presentation name, while \texttt{ExcelOpenFile} requires a valid file path. The argument generator populates these fields using a mixture of rule-based templates, probabilistic sampling from realistic distributions, and generative models generation, ensuring both diversity and plausibility.

\begin{figure}[h]
	\centering
	\includegraphics[width=0.9\linewidth]{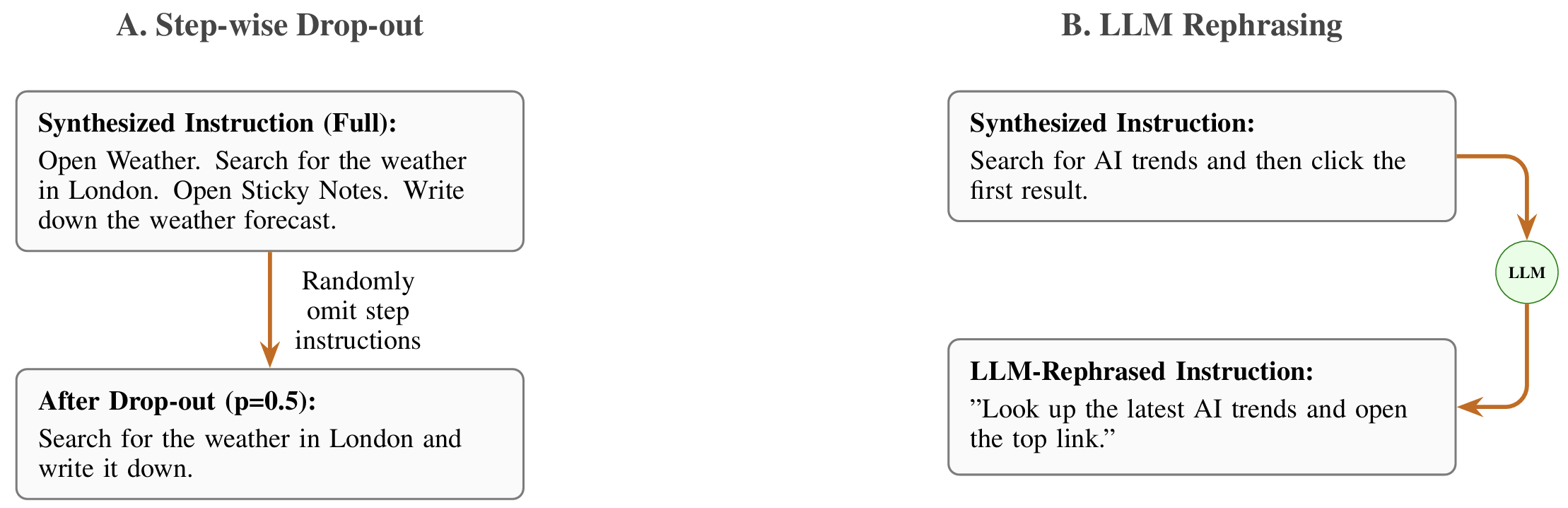}
	\caption{Instruction Narrator with step drop-out and LLM paraphrasing to create realistic user task descriptions.}
	\label{fig:instruction_narrator}
\end{figure}

\paragraph{Instruction Narrator} converts composed atomic task with arguments fulfilled into natural, user-facing instructions. For each composition, the narrator first aggregates atomic task level instructions by randomly sampling from their instruction pools to ensure linguistic diversity. To better reflect real-world user expressions, where users often omit intermediate details, we introduce a step-wise drop-out mechanism that randomly removes a subset of step instructions with a range of probabilities. It produces more concise and human-like directives. Finally, an LLM rephrasing module refines the resulting text, paraphrasing it into fluent, contextually coherent, and user-friendly language. This two-stage realism enhancement, \ie, step-wise drop-out followed by LLM rephrasing, bridges the gap between compositional task generation and realistic natural-language instructions, as illustrated in \autoref{fig:instruction_narrator}.

\section{Experimental Results}

We present a comprehensive evaluation of \algacro{}, encompassing both the process and quality of synthesized user tasks and the performance of different evaluation protocols on application-level tool selection. We begin by examining the user-task generation pipeline and its quality studies in Section~\ref{sec:user_task_generation}, validating the realism and diversity of synthesized instructions. Section~\ref{sec:tool_selection_accuracy} reports results from both baseline and LLM-driven protocols, analyzing overall and per-domain performance across the benchmark. Subsequently, Section~\ref{sec:confusion_analysis} investigates detailed confusion and error analyses that expose systematic weaknesses in inter-application reasoning. Together, these experiments provide an integrated view of data quality and model behavior, establishing a reliable foundation for future research on application-level reasoning and its role in advancing computer-using agents.

\subsection{User Task Generation}\label{sec:user_task_generation}

\paragraph{Experimental Process.}

To generate a sufficiently large, representative, and diverse set of user tasks associated with their target ground-truth applications, we employ the proposed user task generation pipeline described in Section~\ref{sec:user_task_generation}.

In particular, we construct a dataset of atomic tasks spanning the one hundred applications introduced in Section~\ref{sec:application_coverage_and_domains}. The atomic task generation proceeds in two stages. First, we leverage GPT-5~\citep{} with carefully designed system prompts to draft an initial list of atomic tasks. Then, human annotators conduct a thorough refinement process to add missing tasks, remove redundant ones, and improve task clarity. This stage yields several thousand atomic tasks that capture the most common units of everyday user intent. 

Each atomic task may contain one or more parameters requiring realistic argument values. To address this, we develop a set of dedicated argument generators that automatically produce semantically consistent and realistic values. A few representative examples are shown as below.

\begin{tcolorbox}[
  colback=green!10!white,    
  colframe=green!60!black,   
  title=\textbf{Argument Generator Example},
  boxrule=0.4pt,           
  arc=2pt,                 
  left=4pt, right=4pt, top=2pt, bottom=2pt
]
\texttt{generate\_city\_name()}\\
\texttt{generate\_product\_name()}\\
\texttt{generate\_random\_number()}
\end{tcolorbox}

Similarly, we use GPT-5~\citep{} to curate atomic task compositions, which combine multiple atomic tasks into higher-level user activities. Human experts then review and refine these compositions to ensure completeness, diversity, and consistency. By combining these components, we establish a large-scale user task dataset with ground truth application comprising one hundred thousand samples, averaging about one thousand high-quality user tasks per application. To account for tasks that can be accomplished by multiple applications, we further establish an application equivalence mapping that groups functionally overlapping tools under unified task labels. This mapping enables consistent and fair evaluation across interchangeable applications. A few representative user task–application pairs are illustrated below.
\begin{tcolorbox}[
  colback=yellow!10!white,    
  colframe=yellow!60!black,   
  title=\textbf{User Task Application Example (TBD)},
  boxrule=0.4pt,           
  arc=2pt,                 
  left=4pt, right=4pt, top=2pt, bottom=2pt
]
User Task: Send an email to someone.\\
Applications: Outlook\\
User Task: I want to know how is the weather today.\\
Applications: Bing Search, Google Search, Weather, Microsoft Edge, Google Chrome.\\
User Task: Open the cleanup utility and clean drive D:\\
Applications: Disk Cleanup\\
User Task: Navigate to the search menu and search for a PewDiePie Minecraft video.\\
Applications: YouTube\\
User Task: Search for the song 'Shape of You' and play it\\
Applications: Spotify
\end{tcolorbox}

\paragraph{User Task Quality Study.}

To further verify the quality of the generated user task–application pairs, we conducted a series of controlled human evaluations in a laboratory environment. Three complementary experiments were designed to assess different aspects of task quality. Human judges rated each generated user task on a five-point Likert scale (1–5) with respect to \textit{(i) grammatical naturalness}, measuring the grammatical correctness and fluency of the instruction, and \textit{(ii) semantic realism}, evaluating how realistic and plausible each task is in the context of actual computer usage and whether it reflects authentic user intentions rather than artificial phrasing. In parallel, the judges were also asked to assess \textit{(iii) ground-truth application correctness}, verifying whether the annotated target application(s) are appropriate and sufficient to accomplish the described user goal.

Due to the large scale of the dataset, we uniformly sampled 10\% of the total examples—approximately one hundred per application—for human inspection. Each example was independently reviewed by three annotators to ensure consistency and reliability. The evaluation results demonstrate that the generated user tasks are of high quality, achieving an average grammatical score of 4.7, semantic realism of 4.6, and ground-truth correctness of 99.8\%. The high grammatical, semantic, and correctness scores indicate that the generated user task–application pairs closely mirror realistic computer-use scenarios, making the dataset well suited for benchmarking inter-application reasoning.

\begin{table}[t]
\centering
\caption{Accuracy results across \textit{all} application categories.}\label{table:params-all}
\resizebox{0.9\linewidth}{!}{
\begin{tabular}{>{\centering\arraybackslash}p{1.5cm} l >{\centering\arraybackslash}p{1.6cm} >{\centering\arraybackslash}p{1.6cm} >{\centering\arraybackslash}p{1.6cm} c}
\toprule
\multirow{2}{*}{\textbf{Type}} & \multirow{2}{*}{\textbf{Model}} & \multicolumn{3}{c}{\textbf{Method Accuracy}} & \multirow{2}{*}{\textbf{Average}}\\
\cmidrule(lr){3-5}
& & \textbf{Zero-shot} & \textbf{Few-shot} & \textbf{RAS} & \\
\midrule
\multirow{9}{*}{LLM}
& GPT-5~\citep{openai2025gpt5}                   & \cellcolor{blue!10}\textbf{0.620} & \cellcolor{yellow!20}\textbf{0.635} & \cellcolor{green!10}\textbf{0.644} & \textbf{0.633}\\
& GPT-4o-mini~\citep{achiam2023gpt}& \cellcolor{blue!10}0.594 & \cellcolor{yellow!20}0.604 & \cellcolor{green!10}0.611 & 0.603\\
& Qwen-2.5-7B-Instruct~\citep{qwen2024qwen25} & \cellcolor{blue!10}0.530 & \cellcolor{yellow!20}0.550 & \cellcolor{green!10}0.574 & 0.551\\
& Qwen3-4B-Instruct-2507~\citep{yang2025qwen3} & \cellcolor{blue!10}0.521 & \cellcolor{yellow!20}0.529 & \cellcolor{green!10}0.574 & 0.541\\
& Qwen3-30B-A3B-Instruct-2507~\citep{yang2025qwen3} & \cellcolor{blue!10}0.567 & \cellcolor{yellow!20}0.569 & \cellcolor{green!10}0.615 & 0.584\\
& gemma-3-270m~\citep{team2025gemma}& \cellcolor{blue!10}0.024 & \cellcolor{yellow!20}0.244 & \cellcolor{green!10}0.024 & 0.097\\
& gemma-3-4b-pt~\citep{team2025gemma}& \cellcolor{blue!10}0.300 & \cellcolor{yellow!20}0.457 & \cellcolor{green!10}0.370 & 0.376\\
& Lamma-3-8B~\citep{touvron2024llama3}& \cellcolor{blue!10}0.522 & \cellcolor{yellow!20}0.540 & \cellcolor{green!10}0.565 & 0.542\\
& Phi-4~\citep{abdin2024phi}        & \cellcolor{blue!10}{0.504} & \cellcolor{yellow!20}0.540 & \cellcolor{green!10}0.581 & 0.541\\
\midrule
\multirow{2}{*}{Baseline}
& Random Selector                  & \cellcolor{blue!10}-- & \cellcolor{yellow!20}-- & \cellcolor{green!10}-- & 0.016\\
& Rule-based Heuristic             & \cellcolor{blue!10}-- & \cellcolor{yellow!20}-- & \cellcolor{green!10}-- & 0.560\\
\bottomrule
\end{tabular}
}
\end{table}

\subsection{Tool Selection Accuracy}\label{sec:tool_selection_accuracy}

\paragraph{Experimental Setup.}
We evaluate application-level tool selection accuracy across all twelve categories in \textsc{AppSelectBench} under three prompting regimes: \emph{zero-shot}, \emph{few-shot}, and \emph{retrieval-augmented selection (RAS)} and two baselines: \emph{random-selector} and \emph{rule-based heuristic}. Each model receives identical candidate application lists and evaluation instructions (see Appendix~\ref{sec:appendix_prompt}), and inference is conducted with deterministic decoding (\texttt{temperature} = 0) to ensure fair comparison. Accuracy is defined as the proportion of user tasks whose predicted application lies within the annotated set of valid solutions. Because many tasks admit multiple valid tools, correctness is determined by membership rather than exact match to a single canonical label. 

\paragraph{LLM Families.}
To capture both proprietary and open-source behaviors, we evaluate nine representative models spanning distinct architectures and parameter scales:
\begin{itemize}[leftmargin=10pt]
    \item \textbf{Closed-source models:} \textsc{GPT-5}~\citep{openai2025gpt5} and \textsc{GPT-4o-mini}~\citep{achiam2023gpt}, 
    each accessed via API endpoints with consistent decoding parameters.
    \item \textbf{Open-source models:} \textsc{Qwen-2.5-7B-Instruct} and \textsc{Qwen3-30B-A3B-Instruct-2507}~\citep{yang2025qwen3}, 
    \textsc{Llama-3-8B}~\citep{touvron2024llama3}, 
    \textsc{Phi-4}~\citep{abdin2024phi}, and 
    \textsc{Gemma-3} models~\citep{team2025gemma} with 270M and 4B parameters. 
\end{itemize}
This spectrum enables us to analyze how model scale, training data mixture, and instruction tuning affect application-level reasoning.

\begin{figure}[!t]
    \centering
    \includegraphics[width=1.0\linewidth]{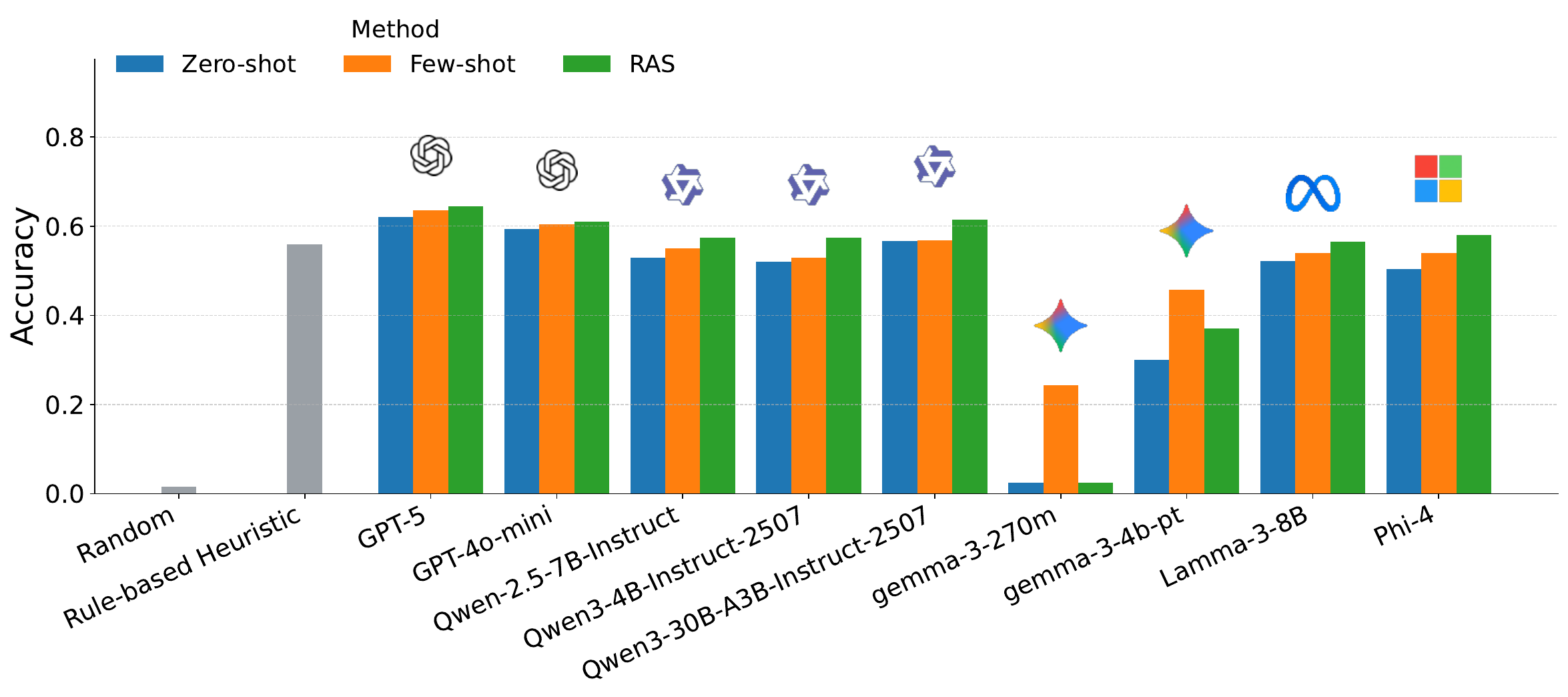}
    \caption{Overall application-selection accuracy for baselines and LLMs under zero-shot, few-shot, and retrieval-augmented selection.}
    \label{fig:grouped_accuracy}
\end{figure}

\paragraph{Baselines and Overall Results.}
\autoref{fig:grouped_accuracy} and \autoref{table:params-all} summarize accuracies across all categories and prompting regimes. 
The \textsc{Random Selector} performs at roughly 1.6\% accuracy, establishing the lower bound, while the \textsc{Rule-based Heuristic} reaches 56\%, confirming that lexical and capability cues explain a substantial subset of straightforward cases yet fail on tasks requiring contextual disambiguation or compositional reasoning.

Among closed-source systems, \textsc{GPT-5} achieves the highest overall accuracy (63.3\%), followed by \textsc{GPT-4o-mini} (60.3\%). 
These results indicate that large proprietary models possess robust internal representations of desktop application semantics, with RAS providing modest stabilization (1–2\%). Open-source models show greater sensitivity to prompting. \textsc{Qwen-2.5-7B-Instruct} improves steadily from zero-shot (53.0\%) $\rightarrow$ few-shot (55.0\%) $\rightarrow$ RAS (57.4\%), closing much of the gap to larger closed models. \textsc{Llama-3-8B} (54.2\%) and \textsc{Phi-4} (54.1\%) achieve comparable mid-tier performance, while extremely small models such as \textsc{Gemma-3-270M} remain unstable, reflecting limited reasoning capacity despite retrieval support.

\paragraph{Effect of Supervision and Retrieval.} Across families, few-shot prompting yields consistent yet moderate gains ($\approx$ +2\%), whereas RAS delivers the largest improvements for mid-scale and open models (+3–5\%). 
For high-capacity models, the benefit of retrieval diminishes as most application semantics are already internalized. 
In contrast, under-parameterized models exhibit non-monotonic behavior, underscoring that retrieval helps only when the model can meaningfully integrate structured descriptions into its reasoning process.

\begin{wrapfigure}{r}{0.6\linewidth} 
  \vspace{-6pt}                        
    \includegraphics[width=\linewidth]{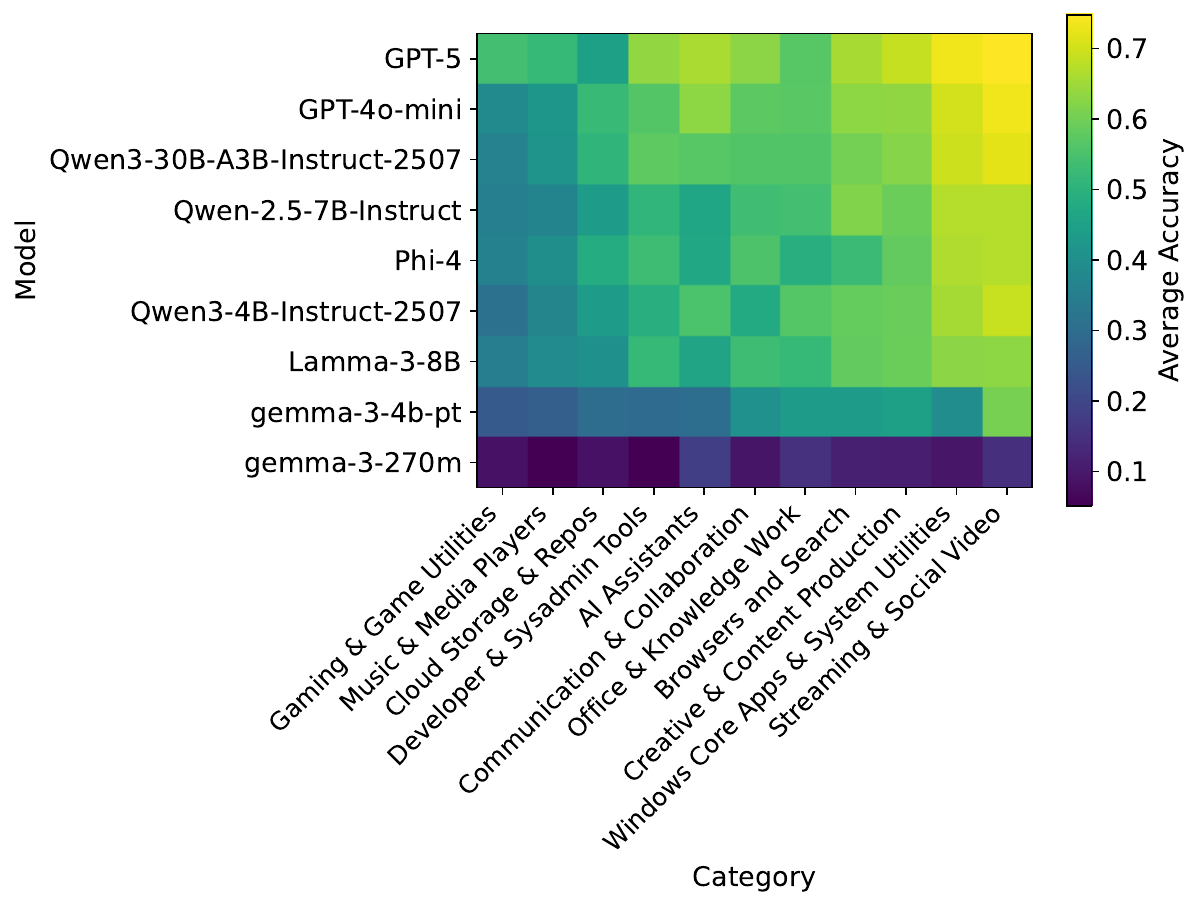}
    \caption{Heatmap of accuracy by model and application category, revealing consistent weaknesses across domains}
    \label{fig:category_model_heatmap}
  \vspace{-11pt}
\end{wrapfigure}
\begin{figure}[!t]
  \centering

  \begin{minipage}[t]{0.55\linewidth}
    \vspace{0pt} 
    \includegraphics[width=\linewidth]{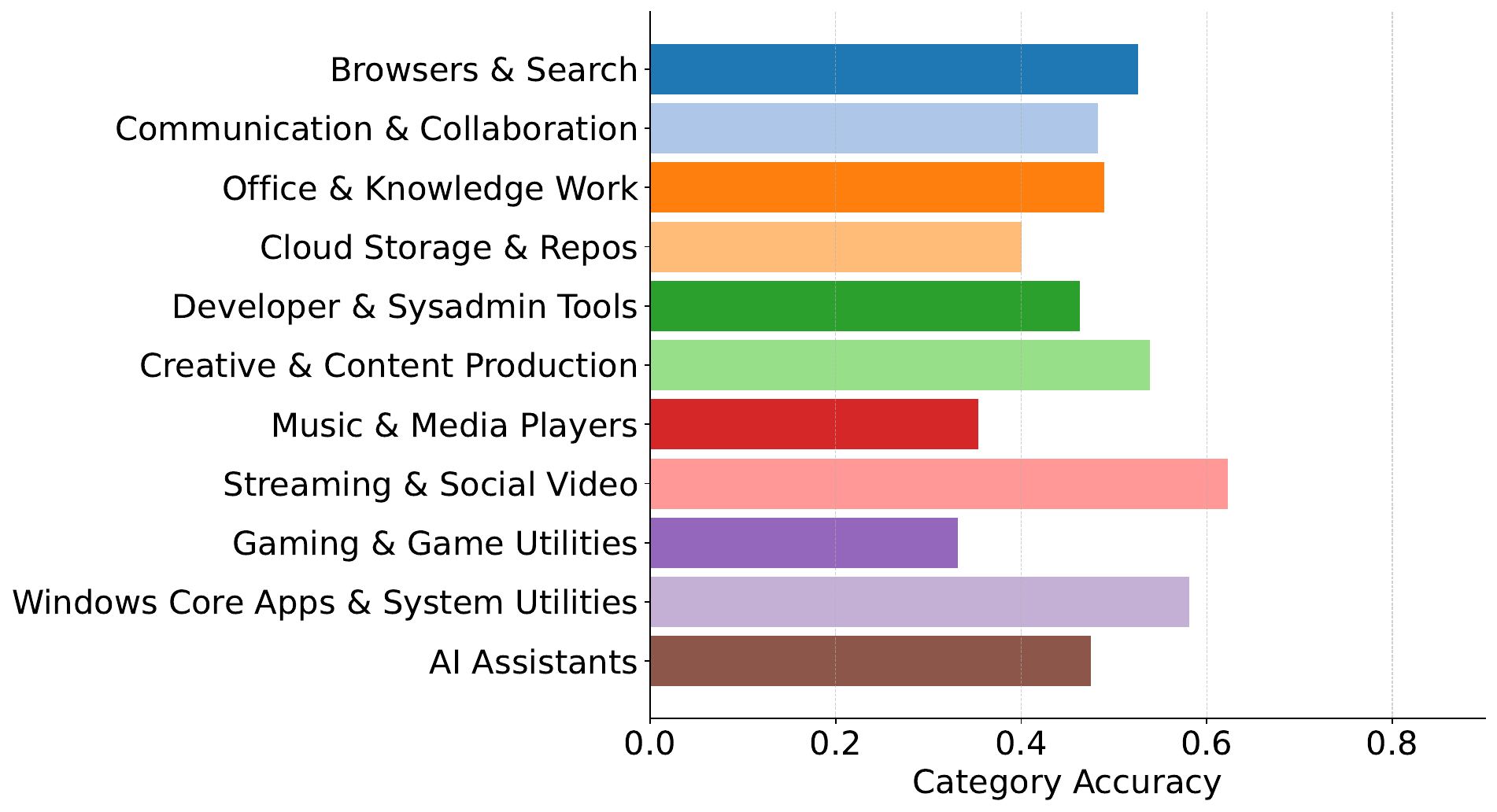}
    \caption{Category-level accuracy, highlighting easier and harder categories with dense functional overlap}
    \label{fig:category_accuracy}
  \end{minipage}
  \hfill
  \begin{minipage}[t]{0.4\linewidth}
    \vspace{0pt} 
    \captionof{table}{Category-level accuracy}
    \label{table:group-accuracy-by-category}
    \resizebox{\linewidth}{!}{
      \begin{tabular}{l c}
        \toprule
        \textbf{Category Type} & \textbf{Category Accuracy} \\
        \midrule
        Streaming \& Social Video & 0.623 \\
        Windows Core Apps \& System Utilities & 0.581 \\
        Creative \& Content Production & 0.539 \\
        Browsers and Search & 0.526 \\
        Office \& Knowledge Work & 0.489 \\
        Communication \& Collaboration & 0.483 \\
        AI Assistants & 0.475 \\
        Developer \& Sysadmin Tools & 0.463 \\
        Cloud Storage \& Repos & 0.401 \\
        Music \& Media Players & 0.354 \\
        Gaming \& Game Utilities & 0.331 \\
        \bottomrule
      \end{tabular}
    }
  \end{minipage}
\end{figure}

\paragraph{Category Performance.} Figure~\ref{fig:category_accuracy} reports category-wise accuracy, computed as the average across the three prompting techniques. We observe a wide spread across categories, indicating that application selection difficulty is highly category-dependent. \textit{Streaming \& Social Video} is the easiest, followed by \textit{Windows Core Apps \& System Utilities}. These categories exhibit specialized behaviors that map to a small set of applications and are well captured by capability descriptions, which makes retrieval-augmented reasoning particularly effective. In contrast, \textit{Gaming \& Game Utilities} and \textit{Music \& Media Players} are hardest. Many tasks involve multiple near-substitutable tools, increasing labeling ambiguity and encouraging confusions among functionally overlapping apps. Mid-tier categories such as 
\textit{Communication \& Collaboration} also contain substantial intra-category overlap, leaving room for improvement in everyday scenarios.

Per-model breakdowns in Figure ~\ref{fig:category_model_heatmap} show that the relative ranking of categories is largely stable across architectures and scales. Stronger models improve performance across all categories but do not eliminate the hardest confusions, while smaller/open models benefit most in categories with distinctive capability profiles. This consistency suggests that much of the variance is dataset and category intrinsic, rather than purely an artifact of a particular model family. Taken together, the category analysis highlights where capability grounding suffices and where richer reasoning is needed.

\subsection{Confusion Analysis}\label{sec:confusion_analysis}

\begin{wrapfigure}{r}{0.50\linewidth} 
  \vspace{-15pt}                       
  \centering
  \includegraphics[width=\linewidth]{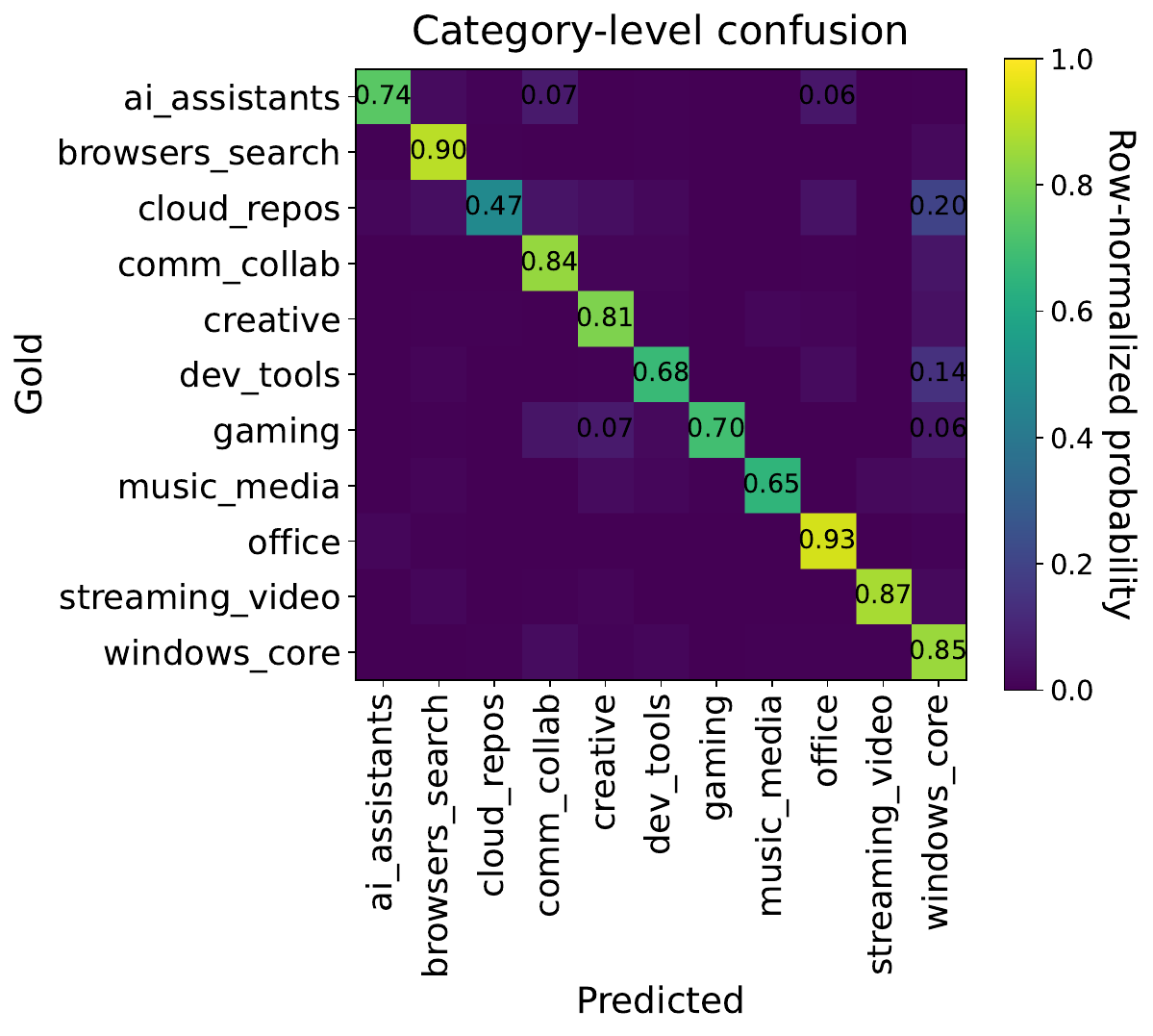}
  \caption{Row-normalized category confusion matrix contrasting systematic cross-category boundary errors}
  \label{fig:category_confusion_analysis}
  \vspace{-11pt}
\end{wrapfigure}
We analyze systematic error patterns using a row-normalized category-level confusion matrix $C \in \mathbb{R}^{K \times K}$, where each entry  $C_{ij}:= \text{Pr}(\mathrm{cat}(\hat{Y}) = j \mid \mathrm{cat}(Y) = i)$ represents the probability of the predicted application $\hat{Y}$ belong to category $j$ given the true category the ground truth application $Y$ falling in category $i$, shown in \autoref{fig:category_confusion_analysis},

The diagonal mass $C_{ii}$ measured per-category cohesion, while off-diagonal entries expose targets that attract misclassifications. Most categories exhibit strong cohesion. \textit{Office \& Knowledge Work} shows the highest diagonal, followed by \textit{Browsers \& Search}. In contrast, \textit{Cloud Storage \& Repos} has comparatively low cohesion with visible leakage into \textit{Windows Core}. This suggests that models often confuse OS-level file operations with cloud sync or coding hosting.

To distinguish between \textbf{intra-category substitutions}, \ie, prediction errors among functionally similar applications within the same category, and \textbf{cross-category boundary errors}, mistakes that cross semantic or functional boundaries between different categories. We decompose misclassifications into two components:
\begin{equation}
\pi_{\text{intra}} = \Pr(\mathrm{cat}(\hat{Y}) = \mathrm{cat}(Y) \mid \hat{Y} \ne Y), 
\qquad 
\pi_{\text{cross}} = 1 - \pi_{\text{intra}}.
\end{equation}
We find $\pi_{\text{intra}} = 0.234$ and $\pi_{\text{cross}} = 0.766$, indicating that cross-category confusions dominate the overall error distribution. In other words, models more often select applications from the wrong functional category (\eg, choosing a cloud storage tool for a file-management task) than confuse similar applications within the same category. This observation suggests that improving category-level discrimination—through hierarchical or modular architectures that first identify the correct category before resolving the specific application—may substantially reduce systematic errors.

To identify consistent, model-independent confusion patterns, we aggregate error pairs $(Y \rightarrow \hat{Y})$ that appear in misclassifications across multiple LLMs. Pairs that recur in at least three distinct models are considered as consistent confusions. These errors are concentrated in categories with overlapping or multi-purpose functionality, \eg, between \textit{Edge} and \textit{Chrome} in web browsing, or between \textit{YouTube} and \textit{Netflix} in media playback. In addition to category-level analysis, we conduct per-application metrics to examine variations across individual applications. Tools with clear and distinctive capabilities, such as Microsoft Word (F1=0.96), exhibit consistently higher precision and recall. Conversely, tools situated within semantically broad or functionally overlapping categories, such as Notepad (F1=0.50), show lower F1 scores due to boundary ambiguity.

\begin{figure}[!hb]
  \centering
  \begin{subfigure}[t]{0.48\linewidth}
    \centering
    \includegraphics[width=\linewidth]{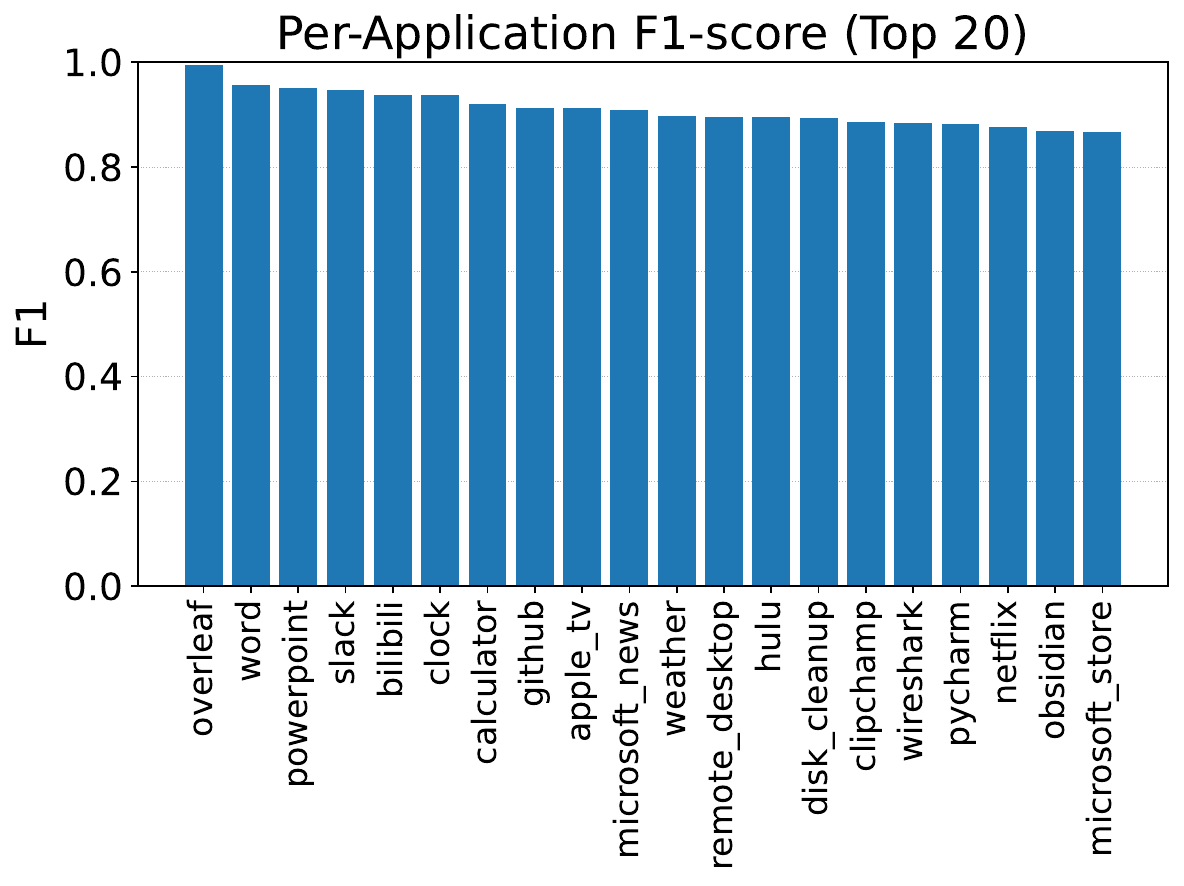}
    \caption{Top F1 score per application}\label{fig:top_f1}
  \end{subfigure}
  \begin{subfigure}[t]{0.48\linewidth}
    \centering
    \includegraphics[width=\linewidth]{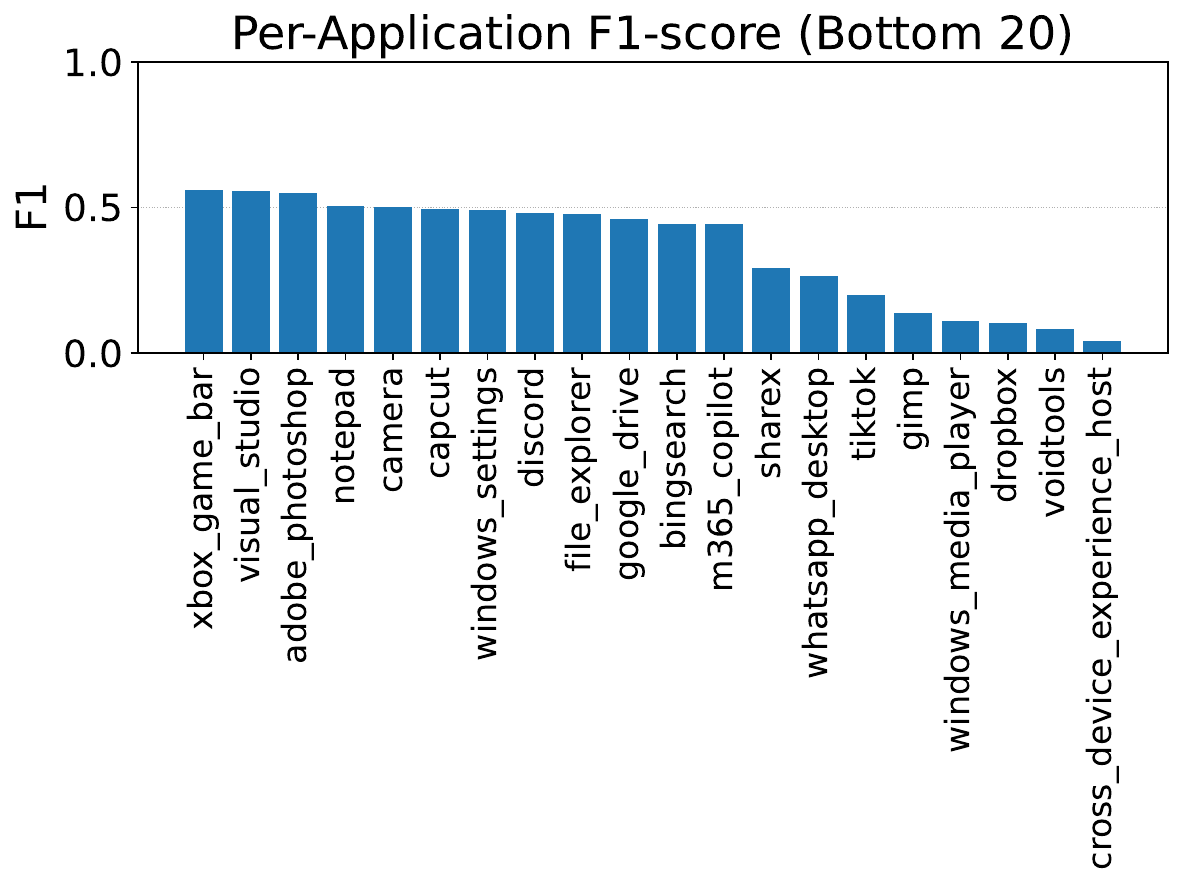}
    \caption{Bottom F1 score per application}\label{fig:bottom_f1}
  \end{subfigure}
  \caption{Top and bottom applications by F1 score}
  \label{fig:f1_side_by_side}
\end{figure}

\section{Conclusion and Future Work}

We introduced \algacro{}, the first comprehensive benchmark for evaluating application-level tool selection in computer-using agents to the best of our knowledge. \algacro{} spans over 100 desktop applications and one thousand realistic user tasks, generated through a novel multi-stage pipeline that composes atomic tasks into fluent natural-language instructions. Together with unified evaluation protocols, the benchmark enables systematic analysis of how large language models map user intents to the correct application environment, bridging the gap between natural-language understanding and executable tool use.

Our experiments show that simple rule-based heuristics relying on keyword or capability matching provides surface-level application-selection. LLMs exhibit broader functional reasoning that generalizes beyond explicit lexical overlap while still has a large room to improve performance. Nevertheless, cross-category confusions remain the dominant source of error, indicating that models often misidentify the functional domain before selecting the correct application.  In future work, we will extend \algacro{} to the multi-application domain, where agents would expect to plan, select, and coordinate multiple tools to complete more complex tasks.

\bibliography{reference}
\bibliographystyle{plain}


\clearpage
\newpage
\appendix
\section{Author List}\label{appendix:authorlist}
\textbf{Microsoft}:\\\\
Tianyi Chen$^{*}$,\\
Michael Solodko$^{*}$, \\ 
Sen Wang, \\
Jongwoo Ko, \\
Junheng Hao, \\
Colby Banbury, \\
Sara Abdali, \\
Saeed Amizadeh, \\
Qing Xiao, \\
Yinheng Li, \\
Tianyu Ding, \\
Kamran Ghasedi Dizaji,\\
Suzhen Zheng, \\
Hao Fan, \\
Justin Wagle, \\
Pashmina Cameron, \\
Kazuhito Koishida,\\

Corresponding Author: Tianyi Chen, \url{Tianyi.Chen@microsoft.com}.\\
Equal Contribution: $^{*}$.

\newpage
\section{Experimental Results}
\captionsetup[table]{skip=8pt}


\begin{table}[ht]
\centering
\caption{Accuracy results (Part I).}
\label{table:all-categories-merged-part1}
\resizebox{0.90\linewidth}{!}{
\begin{tabular}{
    >{\raggedright\arraybackslash}p{3.6cm}
    >{\centering\arraybackslash}p{1.4cm}
    l
    >{\centering\arraybackslash}p{1.5cm}
    >{\centering\arraybackslash}p{1.5cm}
    >{\centering\arraybackslash}p{1.5cm}
    c
}
\toprule
\multirow{2}{*}{\textbf{Category}} & \multirow{2}{*}{\textbf{Type}} & \multirow{2}{*}{\textbf{Model}} & \multicolumn{3}{c}{\textbf{Method Accuracy}} & \multirow{2}{*}{\textbf{Average}}\\
\cmidrule(lr){4-6}
& & & \textbf{Zero-shot} & \textbf{Few-shot} & \textbf{RAS} & \\
\midrule

\multirow{11}{*}{\textit{All Application Domains}}
& \multirow{9}{*}{LLM} & GPT-5 & \cellcolor{blue!10}0.620 & \cellcolor{yellow!20}0.635 & \cellcolor{green!10}0.644 & 0.633 \\
& & GPT-4o-mini & \cellcolor{blue!10}0.594 & \cellcolor{yellow!20}0.604 & \cellcolor{green!10}0.611 & 0.603 \\
& & Qwen-2.5-7B-Instruct & \cellcolor{blue!10}0.530 & \cellcolor{yellow!20}0.550 & \cellcolor{green!10}0.574 & 0.551 \\
& & Qwen3-4B-Instruct-2507 & \cellcolor{blue!10}0.521 & \cellcolor{yellow!20}0.529 & \cellcolor{green!10}0.574 & 0.541 \\
& & Qwen3-30B-A3B-Instruct-2507 & \cellcolor{blue!10}0.567 & \cellcolor{yellow!20}0.569 & \cellcolor{green!10}0.615 & 0.584 \\
& & gemma-3-270m & \cellcolor{blue!10}0.024 & \cellcolor{yellow!20}0.244 & \cellcolor{green!10}0.024 & 0.097 \\
& & gemma-3-4b-pt & \cellcolor{blue!10}0.300 & \cellcolor{yellow!20}0.457 & \cellcolor{green!10}0.370 & 0.376 \\
& & Lamma-3-8B & \cellcolor{blue!10}0.522 & \cellcolor{yellow!20}0.540 & \cellcolor{green!10}0.565 & 0.542 \\
& & Phi-4 & \cellcolor{blue!10}0.504 & \cellcolor{yellow!20}0.540 & \cellcolor{green!10}0.581 & 0.542 \\
\cmidrule(lr){2-7}
& \multirow{2}{*}{Baseline} & Random Selector        & \cellcolor{blue!10}-- & \cellcolor{yellow!20}-- & \cellcolor{green!10}-- & 0.016 \\
& & Rule-based Heuristic  & \cellcolor{blue!10}-- & \cellcolor{yellow!20}-- & \cellcolor{green!10}-- & 0.560 \\
\midrule

\multirow{11}{*}{\textit{Browsers \& Search}}
& \multirow{9}{*}{LLM} & GPT-5 & \cellcolor{blue!10}0.657 & \cellcolor{yellow!20}0.668 & \cellcolor{green!10}0.643 & 0.656 \\
& & GPT-4o-mini & \cellcolor{blue!10}0.629 & \cellcolor{yellow!20}0.624 & \cellcolor{green!10}0.628 & 0.628 \\
& & Qwen-2.5-7B-Instruct & \cellcolor{blue!10}0.615 & \cellcolor{yellow!20}0.613 & \cellcolor{green!10}0.619 & 0.615 \\
& & Qwen3-4B-Instruct-2507 & \cellcolor{blue!10}0.576 & \cellcolor{yellow!20}0.567 & \cellcolor{green!10}0.610 & 0.584 \\
& & Qwen3-30B-A3B-Instruct-2507 & \cellcolor{blue!10}0.569 & \cellcolor{yellow!20}0.622 & \cellcolor{green!10}0.614 & 0.602 \\
& & gemma-3-270m & \cellcolor{blue!10}0.026 & \cellcolor{yellow!20}0.292 & \cellcolor{green!10}0.015 & 0.111 \\
& & gemma-3-4b-pt & \cellcolor{blue!10}0.396 & \cellcolor{yellow!20}0.564 & \cellcolor{green!10}0.347 & 0.436 \\
& & Lamma-3-8B & \cellcolor{blue!10}0.555 & \cellcolor{yellow!20}0.598 & \cellcolor{green!10}0.586 & 0.580 \\
& & Phi-4 & \cellcolor{blue!10}0.431 & \cellcolor{yellow!20}0.543 & \cellcolor{green!10}0.595 & 0.523 \\
\cmidrule(lr){2-7}
& \multirow{2}{*}{Baseline} & Random Selector        & \cellcolor{blue!10}-- & \cellcolor{yellow!20}-- & \cellcolor{green!10}-- & 0.009 \\
& & Rule-based Heuristic  & \cellcolor{blue!10}-- & \cellcolor{yellow!20}-- & \cellcolor{green!10}-- & 0.609 \\
\midrule

\multirow{11}{*}{\textit{Communication \& Collaboration}}
& \multirow{9}{*}{LLM} & GPT-5 & \cellcolor{blue!10}0.616 & \cellcolor{yellow!20}0.639 & \cellcolor{green!10}0.626 & 0.627 \\
& & GPT-4o-mini & \cellcolor{blue!10}0.585 & \cellcolor{yellow!20}0.559 & \cellcolor{green!10}0.569 & 0.571 \\
& & Qwen-2.5-7B-Instruct & \cellcolor{blue!10}0.543 & \cellcolor{yellow!20}0.531 & \cellcolor{green!10}0.530 & 0.535 \\
& & Qwen3-4B-Instruct-2507 & \cellcolor{blue!10}0.477 & \cellcolor{yellow!20}0.442 & \cellcolor{green!10}0.510 & 0.476 \\
& & Qwen3-30B-A3B-Instruct-2507 & \cellcolor{blue!10}0.546 & \cellcolor{yellow!20}0.558 & \cellcolor{green!10}0.573 & 0.559 \\
& & gemma-3-270m & \cellcolor{blue!10}0.028 & \cellcolor{yellow!20}0.223 & \cellcolor{green!10}0.023 & 0.091 \\
& & gemma-3-4b-pt & \cellcolor{blue!10}0.338 & \cellcolor{yellow!20}0.490 & \cellcolor{green!10}0.383 & 0.404 \\
& & Lamma-3-8B & \cellcolor{blue!10}0.523 & \cellcolor{yellow!20}0.567 & \cellcolor{green!10}0.502 & 0.531 \\
& & Phi-4 & \cellcolor{blue!10}0.553 & \cellcolor{yellow!20}0.572 & \cellcolor{green!10}0.536 & 0.554 \\
\cmidrule(lr){2-7}
& \multirow{2}{*}{Baseline} & Random Selector        & \cellcolor{blue!10}-- & \cellcolor{yellow!20}-- & \cellcolor{green!10}-- & 0.010 \\
& & Rule-based Heuristic  & \cellcolor{blue!10}-- & \cellcolor{yellow!20}-- & \cellcolor{green!10}-- & 0.523 \\
\midrule

\multirow{11}{*}{\textit{Office \& Knowledge Work}}
& \multirow{9}{*}{LLM} & GPT-5 & \cellcolor{blue!10}0.560 & \cellcolor{yellow!20}0.571 & \cellcolor{green!10}0.574 & 0.568 \\
& & GPT-4o-mini & \cellcolor{blue!10}0.570 & \cellcolor{yellow!20}0.577 & \cellcolor{green!10}0.565 & 0.571 \\
& & Qwen-2.5-7B-Instruct & \cellcolor{blue!10}0.550 & \cellcolor{yellow!20}0.557 & \cellcolor{green!10}0.514 & 0.540 \\
& & Qwen3-4B-Instruct-2507 & \cellcolor{blue!10}0.561 & \cellcolor{yellow!20}0.569 & \cellcolor{green!10}0.563 & 0.564 \\
& & Qwen3-30B-A3B-Instruct-2507 & \cellcolor{blue!10}0.564 & \cellcolor{yellow!20}0.571 & \cellcolor{green!10}0.545 & 0.560 \\
& & gemma-3-270m & \cellcolor{blue!10}0.022 & \cellcolor{yellow!20}0.415 & \cellcolor{green!10}0.012 & 0.150 \\
& & gemma-3-4b-pt & \cellcolor{blue!10}0.347 & \cellcolor{yellow!20}0.536 & \cellcolor{green!10}0.419 & 0.434 \\
& & Lamma-3-8B & \cellcolor{blue!10}0.479 & \cellcolor{yellow!20}0.522 & \cellcolor{green!10}0.555 & 0.519 \\
& & Phi-4 & \cellcolor{blue!10}0.427 & \cellcolor{yellow!20}0.529 & \cellcolor{green!10}0.555 & 0.503 \\
\cmidrule(lr){2-7}
& \multirow{2}{*}{Baseline} & Random Selector        & \cellcolor{blue!10}-- & \cellcolor{yellow!20}-- & \cellcolor{green!10}-- & 0.008 \\
& & Rule-based Heuristic  & \cellcolor{blue!10}-- & \cellcolor{yellow!20}-- & \cellcolor{green!10}-- & 0.556 \\
\bottomrule
\end{tabular}
}
\end{table}

\begin{table}[ht]
\centering
\caption{Accuracy results (Part II).}
\label{table:all-categories-merged-part2}
\resizebox{0.90\linewidth}{!}{
\begin{tabular}{
    >{\raggedright\arraybackslash}p{3.6cm}
    >{\centering\arraybackslash}p{1.4cm}
    l
    >{\centering\arraybackslash}p{1.5cm}
    >{\centering\arraybackslash}p{1.5cm}
    >{\centering\arraybackslash}p{1.5cm}
    c
}
\toprule
\multirow{2}{*}{\textbf{Category}} & \multirow{2}{*}{\textbf{Type}} & \multirow{2}{*}{\textbf{Model}} & \multicolumn{3}{c}{\textbf{Method Accuracy}} & \multirow{2}{*}{\textbf{Average}}\\
\cmidrule(lr){4-6}
& & & \textbf{Zero-shot} & \textbf{Few-shot} & \textbf{RAS} & \\
\midrule

\multirow{11}{*}{\textit{Cloud Storage \& Repos}}
& \multirow{9}{*}{LLM} & GPT-5 & \cellcolor{blue!10}0.411 & \cellcolor{yellow!20}0.491 & \cellcolor{green!10}0.438 & 0.446 \\
& & GPT-4o-mini & \cellcolor{blue!10}0.516 & \cellcolor{yellow!20}0.531 & \cellcolor{green!10}0.516 & 0.518 \\
& & Qwen-2.5-7B-Instruct & \cellcolor{blue!10}0.389 & \cellcolor{yellow!20}0.434 & \cellcolor{green!10}0.476 & 0.433 \\
& & Qwen3-4B-Instruct-2507 & \cellcolor{blue!10}0.382 & \cellcolor{yellow!20}0.486 & \cellcolor{green!10}0.435 & 0.434 \\
& & Qwen3-30B-A3B-Instruct-2507 & \cellcolor{blue!10}0.492 & \cellcolor{yellow!20}0.494 & \cellcolor{green!10}0.531 & 0.505 \\
& & gemma-3-270m & \cellcolor{blue!10}0.021 & \cellcolor{yellow!20}0.219 & \cellcolor{green!10}0.016 & 0.085 \\
& & gemma-3-4b-pt & \cellcolor{blue!10}0.164 & \cellcolor{yellow!20}0.436 & \cellcolor{green!10}0.297 & 0.299 \\
& & Lamma-3-8B & \cellcolor{blue!10}0.367 & \cellcolor{yellow!20}0.457 & \cellcolor{green!10}0.377 & 0.400 \\
& & Phi-4 & \cellcolor{blue!10}0.467 & \cellcolor{yellow!20}0.516 & \cellcolor{green!10}0.466 & 0.483 \\
\cmidrule(lr){2-7}
& \multirow{2}{*}{Baseline} & Random Selector        & \cellcolor{blue!10}-- & \cellcolor{yellow!20}-- & \cellcolor{green!10}-- & 0.004 \\
& & Rule-based Heuristic  & \cellcolor{blue!10}-- & \cellcolor{yellow!20}-- & \cellcolor{green!10}-- & 0.439 \\
\midrule

\multirow{11}{*}{\textit{Developer \& Sysadmin Tools}}
& \multirow{9}{*}{LLM} & GPT-5 & \cellcolor{blue!10}0.617 & \cellcolor{yellow!20}0.636 & \cellcolor{green!10}0.655 & 0.636 \\
& & GPT-4o-mini & \cellcolor{blue!10}0.544 & \cellcolor{yellow!20}0.556 & \cellcolor{green!10}0.586 & 0.561 \\
& & Qwen-2.5-7B-Instruct & \cellcolor{blue!10}0.448 & \cellcolor{yellow!20}0.494 & \cellcolor{green!10}0.585 & 0.509 \\
& & Qwen3-4B-Instruct-2507 & \cellcolor{blue!10}0.427 & \cellcolor{yellow!20}0.453 & \cellcolor{green!10}0.588 & 0.489 \\
& & Qwen3-30B-A3B-Instruct-2507 & \cellcolor{blue!10}0.545 & \cellcolor{yellow!20}0.558 & \cellcolor{green!10}0.620 & 0.574 \\
& & gemma-3-270m & \cellcolor{blue!10}0.030 & \cellcolor{yellow!20}0.098 & \cellcolor{green!10}0.032 & 0.053 \\
& & gemma-3-4b-pt & \cellcolor{blue!10}0.219 & \cellcolor{yellow!20}0.381 & \cellcolor{green!10}0.282 & 0.294 \\
& & Lamma-3-8B & \cellcolor{blue!10}0.456 & \cellcolor{yellow!20}0.497 & \cellcolor{green!10}0.597 & 0.516 \\
& & Phi-4 & \cellcolor{blue!10}0.485 & \cellcolor{yellow!20}0.508 & \cellcolor{green!10}0.602 & 0.531 \\
\cmidrule(lr){2-7}
& \multirow{2}{*}{Baseline} & Random Selector        & \cellcolor{blue!10}-- & \cellcolor{yellow!20}-- & \cellcolor{green!10}-- & 0.007 \\
& & Rule-based Heuristic  & \cellcolor{blue!10}-- & \cellcolor{yellow!20}-- & \cellcolor{green!10}-- & 0.569 \\
\midrule

\multirow{11}{*}{\textit{Creative \& Content Production}}
& \multirow{9}{*}{LLM} & GPT-5 & \cellcolor{blue!10}0.670 & \cellcolor{yellow!20}0.698 & \cellcolor{green!10}0.695 & 0.687 \\
& & GPT-4o-mini & \cellcolor{blue!10}0.626 & \cellcolor{yellow!20}0.634 & \cellcolor{green!10}0.640 & 0.633 \\
& & Qwen-2.5-7B-Instruct & \cellcolor{blue!10}0.593 & \cellcolor{yellow!20}0.616 & \cellcolor{green!10}0.568 & 0.592 \\
& & Qwen3-4B-Instruct-2507 & \cellcolor{blue!10}0.559 & \cellcolor{yellow!20}0.577 & \cellcolor{green!10}0.630 & 0.588 \\
& & Qwen3-30B-A3B-Instruct-2507 & \cellcolor{blue!10}0.596 & \cellcolor{yellow!20}0.607 & \cellcolor{green!10}0.663 & 0.622 \\
& & gemma-3-270m & \cellcolor{blue!10}0.033 & \cellcolor{yellow!20}0.281 & \cellcolor{green!10}0.017 & 0.110 \\
& & gemma-3-4b-pt & \cellcolor{blue!10}0.379 & \cellcolor{yellow!20}0.530 & \cellcolor{green!10}0.432 & 0.447 \\
& & Lamma-3-8B & \cellcolor{blue!10}0.580 & \cellcolor{yellow!20}0.595 & \cellcolor{green!10}0.595 & 0.590 \\
& & Phi-4 & \cellcolor{blue!10}0.521 & \cellcolor{yellow!20}0.603 & \cellcolor{green!10}0.615 & 0.579 \\
\cmidrule(lr){2-7}
& \multirow{2}{*}{Baseline} & Random Selector        & \cellcolor{blue!10}-- & \cellcolor{yellow!20}-- & \cellcolor{green!10}-- & 0.016 \\
& & Rule-based Heuristic  & \cellcolor{blue!10}-- & \cellcolor{yellow!20}-- & \cellcolor{green!10}-- & 0.616 \\
\midrule

\multirow{11}{*}{\textit{Music \& Media Players}}
& \multirow{9}{*}{LLM} & GPT-5 & \cellcolor{blue!10}0.511 & \cellcolor{yellow!20}0.534 & \cellcolor{green!10}0.509 & 0.518 \\
& & GPT-4o-mini & \cellcolor{blue!10}0.412 & \cellcolor{yellow!20}0.435 & \cellcolor{green!10}0.416 & 0.421 \\
& & Qwen-2.5-7B-Instruct & \cellcolor{blue!10}0.337 & \cellcolor{yellow!20}0.389 & \cellcolor{green!10}0.371 & 0.365 \\
& & Qwen3-4B-Instruct-2507 & \cellcolor{blue!10}0.344 & \cellcolor{yellow!20}0.382 & \cellcolor{green!10}0.374 & 0.366 \\
& & Qwen3-30B-A3B-Instruct-2507 & \cellcolor{blue!10}0.418 & \cellcolor{yellow!20}0.417 & \cellcolor{green!10}0.400 & 0.411 \\
& & gemma-3-270m & \cellcolor{blue!10}0.003 & \cellcolor{yellow!20}0.142 & \cellcolor{green!10}0.007 & 0.050 \\
& & gemma-3-4b-pt & \cellcolor{blue!10}0.235 & \cellcolor{yellow!20}0.238 & \cellcolor{green!10}0.316 & 0.263 \\
& & Lamma-3-8B & \cellcolor{blue!10}0.379 & \cellcolor{yellow!20}0.382 & \cellcolor{green!10}0.401 & 0.387 \\
& & Phi-4 & \cellcolor{blue!10}0.399 & \cellcolor{yellow!20}0.399 & \cellcolor{green!10}0.400 & 0.399 \\
\cmidrule(lr){2-7}
& \multirow{2}{*}{Baseline} & Random Selector        & \cellcolor{blue!10}-- & \cellcolor{yellow!20}-- & \cellcolor{green!10}-- & 0.002 \\
& & Rule-based Heuristic  & \cellcolor{blue!10}-- & \cellcolor{yellow!20}-- & \cellcolor{green!10}-- & 0.399 \\
\bottomrule
\end{tabular}
}
\end{table}

\begin{table}[ht]
\centering
\caption{Accuracy results (Part III).}
\label{table:all-categories-merged-part3}
\resizebox{0.90\linewidth}{!}{
\begin{tabular}{
    >{\raggedright\arraybackslash}p{3.6cm}
    >{\centering\arraybackslash}p{1.4cm}
    l
    >{\centering\arraybackslash}p{1.5cm}
    >{\centering\arraybackslash}p{1.5cm}
    >{\centering\arraybackslash}p{1.5cm}
    c
}
\toprule
\multirow{2}{*}{\textbf{Category}} & \multirow{2}{*}{\textbf{Type}} & \multirow{2}{*}{\textbf{Model}} & \multicolumn{3}{c}{\textbf{Method Accuracy}} & \multirow{2}{*}{\textbf{Average}}\\
\cmidrule(lr){4-6}
& & & \textbf{Zero-shot} & \textbf{Few-shot} & \textbf{RAS} & \\
\midrule

\multirow{11}{*}{\textit{Streaming \& Social Video}}
& \multirow{9}{*}{LLM} & GPT-5 & \cellcolor{blue!10}0.733 & \cellcolor{yellow!20}0.742 & \cellcolor{green!10}0.768 & 0.747 \\
& & GPT-4o-mini & \cellcolor{blue!10}0.711 & \cellcolor{yellow!20}0.721 & \cellcolor{green!10}0.766 & 0.732 \\
& & Qwen-2.5-7B-Instruct & \cellcolor{blue!10}0.658 & \cellcolor{yellow!20}0.675 & \cellcolor{green!10}0.680 & 0.671 \\
& & Qwen3-4B-Instruct-2507 & \cellcolor{blue!10}0.672 & \cellcolor{yellow!20}0.673 & \cellcolor{green!10}0.720 & 0.688 \\
& & Qwen3-30B-A3B-Instruct-2507 & \cellcolor{blue!10}0.702 & \cellcolor{yellow!20}0.707 & \cellcolor{green!10}0.749 & 0.719 \\
& & gemma-3-270m & \cellcolor{blue!10}0.028 & \cellcolor{yellow!20}0.382 & \cellcolor{green!10}0.026 & 0.145 \\
& & gemma-3-4b-pt & \cellcolor{blue!10}0.545 & \cellcolor{yellow!20}0.628 & \cellcolor{green!10}0.640 & 0.604 \\
& & Lamma-3-8B & \cellcolor{blue!10}0.577 & \cellcolor{yellow!20}0.672 & \cellcolor{green!10}0.634 & 0.627 \\
& & Phi-4 & \cellcolor{blue!10}0.644 & \cellcolor{yellow!20}0.676 & \cellcolor{green!10}0.688 & 0.669 \\
\cmidrule(lr){2-7}
& \multirow{2}{*}{Baseline} & Random Selector        & \cellcolor{blue!10}-- & \cellcolor{yellow!20}-- & \cellcolor{green!10}-- & 0.013 \\
& & Rule-based Heuristic  & \cellcolor{blue!10}-- & \cellcolor{yellow!20}-- & \cellcolor{green!10}-- & 0.754 \\
\midrule

\multirow{11}{*}{\textit{Gaming \& Game Utilities}}
& \multirow{9}{*}{LLM} & GPT-5 & \cellcolor{blue!10}0.535 & \cellcolor{yellow!20}0.542 & \cellcolor{green!10}0.538 & 0.538 \\
& & GPT-4o-mini & \cellcolor{blue!10}0.369 & \cellcolor{yellow!20}0.369 & \cellcolor{green!10}0.410 & 0.382 \\
& & Qwen-2.5-7B-Instruct & \cellcolor{blue!10}0.346 & \cellcolor{yellow!20}0.317 & \cellcolor{green!10}0.390 & 0.351 \\
& & Qwen3-4B-Instruct-2507 & \cellcolor{blue!10}0.291 & \cellcolor{yellow!20}0.306 & \cellcolor{green!10}0.343 & 0.313 \\
& & Qwen3-30B-A3B-Instruct-2507 & \cellcolor{blue!10}0.354 & \cellcolor{yellow!20}0.330 & \cellcolor{green!10}0.395 & 0.359 \\
& & gemma-3-270m & \cellcolor{blue!10}0.069 & \cellcolor{yellow!20}0.172 & \cellcolor{green!10}0.018 & 0.086 \\
& & gemma-3-4b-pt & \cellcolor{blue!10}0.183 & \cellcolor{yellow!20}0.298 & \cellcolor{green!10}0.261 & 0.247 \\
& & Lamma-3-8B & \cellcolor{blue!10}0.337 & \cellcolor{yellow!20}0.350 & \cellcolor{green!10}0.359 & 0.348 \\
& & Phi-4 & \cellcolor{blue!10}0.328 & \cellcolor{yellow!20}0.335 & \cellcolor{green!10}0.404 & 0.355 \\
\cmidrule(lr){2-7}
& \multirow{2}{*}{Baseline} & Random Selector        & \cellcolor{blue!10}-- & \cellcolor{yellow!20}-- & \cellcolor{green!10}-- & 0.005 \\
& & Rule-based Heuristic  & \cellcolor{blue!10}-- & \cellcolor{yellow!20}-- & \cellcolor{green!10}-- & 0.392 \\
\midrule

\multirow{11}{*}{\textit{Windows Core Apps \& System Utilities}}
& \multirow{9}{*}{LLM} & GPT-5 & \cellcolor{blue!10}0.724 & \cellcolor{yellow!20}0.741 & \cellcolor{green!10}0.729 & 0.731 \\
& & GPT-4o-mini & \cellcolor{blue!10}0.691 & \cellcolor{yellow!20}0.707 & \cellcolor{green!10}0.700 & 0.699 \\
& & Qwen-2.5-7B-Instruct & \cellcolor{blue!10}0.645 & \cellcolor{yellow!20}0.686 & \cellcolor{green!10}0.678 & 0.669 \\
& & Qwen3-4B-Instruct-2507 & \cellcolor{blue!10}0.599 & \cellcolor{yellow!20}0.682 & \cellcolor{green!10}0.679 & 0.653 \\
& & Qwen3-30B-A3B-Instruct-2507 & \cellcolor{blue!10}0.688 & \cellcolor{yellow!20}0.698 & \cellcolor{green!10}0.700 & 0.695 \\
& & gemma-3-270m & \cellcolor{blue!10}0.010 & \cellcolor{yellow!20}0.256 & \cellcolor{green!10}0.017 & 0.094 \\
& & gemma-3-4b-pt & \cellcolor{blue!10}0.330 & \cellcolor{yellow!20}0.550 & \cellcolor{green!10}0.298 & 0.392 \\
& & Lamma-3-8B & \cellcolor{blue!10}0.612 & \cellcolor{yellow!20}0.610 & \cellcolor{green!10}0.660 & 0.627 \\
& & Phi-4 & \cellcolor{blue!10}0.652 & \cellcolor{yellow!20}0.682 & \cellcolor{green!10}0.657 & 0.663 \\
\cmidrule(lr){2-7}
& \multirow{2}{*}{Baseline} & Random Selector        & \cellcolor{blue!10}-- & \cellcolor{yellow!20}-- & \cellcolor{green!10}-- & 0.007 \\
& & Rule-based Heuristic  & \cellcolor{blue!10}-- & \cellcolor{yellow!20}-- & \cellcolor{green!10}-- & 0.630 \\
\midrule

\multirow{11}{*}{\textit{AI Assistants}}
& \multirow{9}{*}{LLM} & GPT-5 & \cellcolor{blue!10}0.642 & \cellcolor{yellow!20}0.685 & \cellcolor{green!10}0.647 & 0.658 \\
& & GPT-4o-mini & \cellcolor{blue!10}0.594 & \cellcolor{yellow!20}0.615 & \cellcolor{green!10}0.677 & 0.628 \\
& & Qwen-2.5-7B-Instruct & \cellcolor{blue!10}0.435 & \cellcolor{yellow!20}0.465 & \cellcolor{green!10}0.489 & 0.463 \\
& & Qwen3-4B-Instruct-2507 & \cellcolor{blue!10}0.554 & \cellcolor{yellow!20}0.552 & \cellcolor{green!10}0.540 & 0.548 \\
& & Qwen3-30B-A3B-Instruct-2507 & \cellcolor{blue!10}0.558 & \cellcolor{yellow!20}0.577 & \cellcolor{green!10}0.565 & 0.566 \\
& & gemma-3-270m & \cellcolor{blue!10}0.122 & \cellcolor{yellow!20}0.311 & \cellcolor{green!10}0.103 & 0.178 \\
& & gemma-3-4b-pt & \cellcolor{blue!10}0.156 & \cellcolor{yellow!20}0.185 & \cellcolor{green!10}0.567 & 0.302 \\
& & Lamma-3-8B & \cellcolor{blue!10}0.451 & \cellcolor{yellow!20}0.448 & \cellcolor{green!10}0.474 & 0.457 \\
& & Phi-4 & \cellcolor{blue!10}0.422 & \cellcolor{yellow!20}0.485 & \cellcolor{green!10}0.497 & 0.468 \\
\cmidrule(lr){2-7}
& \multirow{2}{*}{Baseline} & Random Selector        & \cellcolor{blue!10}-- & \cellcolor{yellow!20}-- & \cellcolor{green!10}-- & 0.010 \\
& & Rule-based Heuristic  & \cellcolor{blue!10}-- & \cellcolor{yellow!20}-- & \cellcolor{green!10}-- & 0.668 \\
\bottomrule
\end{tabular}
}
\end{table}

\clearpage
\section{Evaluation System Prompt}\label{sec:appendix_prompt}
\captionsetup[table]{skip=8pt}

\begin{tcolorbox}[
  colback=blue!10!white,    
  colframe=blue!60!black,   
  title=\textbf{LLM System Prompt},
  boxrule=0.4pt,            
  arc=2pt,                  
  left=4pt, right=4pt, top=2pt, bottom=2pt,
  fonttitle=\bfseries,
]
\small
You are an expert assistant that selects the best application for accomplishing a task. 
Respond with only the application name. 

\textbf{Available applications:} {7zip, adobe\_acrobat\_reader, adobe\_photoshop, amazon, android\_studio, app\_installer, apple\_tv, battle.net, bilibili, bing\_maps, bing\_search, blender, bluetooth\_settings, calculator, camera, canva, capcut, chrome, clipchamp, clock, command\_prompt, control\_panel, cross\_device\_experience\_host, device\_manager, discord, disk\_cleanup, docker\_desktop, dropbox, excel, file\_explorer, get\_help, gimp, github, google\_drive, groove\_music, handbrake, hulu, intellij\_idea, itunes, m365\_copilot, matlab, microsoft\_copilot, microsoft\_edge, microsoft\_news, microsoft\_onenote, microsoft\_powertoys, microsoft\_store, microsoft\_teams, microsoft\_to\_do, msexpense, netease\_cloud\_music\_, netflix, notepad, notepad++, notion, obs\_studio, obsidian, onedrive, outlook, overleaf, paint, phone\_link, photos, power\_automate, powerpoint, powershell, prime\_video, putty, pycharm, quick\_assist, remote\_desktop, rstudio, rufus, sharex, slack, snipping\_tool, solitaire\_collection, sound\_recorder, spotify, steam, sticky\_notes, task\_manager, tiktok, visual\_studio, visual\_studio\_code, vlc\_media\_player, voidtools, weather, whatsapp\_desktop, windows\_media\_player, windows\_security, windows\_settings, windows\_subsystem\_for\_linux\_(wsl), wireshark, word, xbox\_game\_bar, xbox\_speech-to-text\_overlay, youtube, zoom.}
\end{tcolorbox}
\clearpage
\section{Application Domains}\label{sec:application_domain}
\captionsetup[table]{skip=8pt} 

\begin{table}[ht]
\centering
\small
\begin{tabularx}{\textwidth}{>{\raggedright\arraybackslash}p{0.30\textwidth} >{\raggedright\arraybackslash}X}
\toprule
\textbf{Application Category} & \textbf{Applications} \\
\midrule
Browsers \& Search & Microsoft Edge, Google Chrome, Bing Search, Amazon\\
\midrule
Communication \& Collaboration & Microsoft Teams, Slack, Zoom, Discord, WhatsApp Desktop, Outlook, Remote Desktop, Phone Link \\
\midrule
Office \& Knowledge Work & Word, Excel, PowerPoint, Microsoft OneNote, Notion, Obsidian, Overleaf, Sticky Notes, Microsoft To Do, Power Automate, Adobe Acrobat Reader, MSExpense \\
\midrule
Cloud Storage \& Repos & OneDrive, Google Drive, Dropbox, GitHub \\
\midrule
Developer \& Sysadmin Tools & \textit{IDEs \& editors:} Visual Studio Code, Visual Studio, PyCharm, IntelliJ IDEA, Android Studio, RStudio, MATLAB, Notepad++; \newline
\textit{Shells \& runtimes:} PowerShell, Command Prompt, Windows Subsystem for Linux (WSL); \newline
\textit{Containers:} Docker Desktop; \newline
\textit{Networking/remote tooling:} Wireshark, PuTTY; \newline
\textit{Utilities/build/provisioning:} 7-Zip, Rufus, Voidtools, Microsoft PowerToys, App Installer \\
\midrule
Creative \& Content Production & Adobe Photoshop, GIMP, Canva, Blender, CapCut, Clipchamp, OBS Studio, ShareX, Snipping Tool, Photos, Paint, Camera, Sound Recorder, HandBrake, Word, Excel, PowerPoint \\
\midrule
Music \& Media Players & VLC Media Player, Windows Media Player, Groove Music, Spotify, iTunes, NetEase Cloud Music \\
\midrule
Streaming \& Social Video & YouTube, Netflix, Amazon Prime Video, AppleTV, Hulu, Bilibili, TikTok \\
\midrule
Gaming \& Game Utilities & Steam, Battle.net, Xbox Game Bar, Xbox Speech-to-Text Overlay, Solitaire Collection \\
\midrule
Windows Core Apps \& System Utilities & File Explorer, Windows Settings, Control Panel, Device Manager, Task Manager, Disk Cleanup, Bluetooth Settings, Microsoft Store, Get Help, Quick Assist, Cross Device Experience Host, Windows Security, Weather, Clock, Calculator, Bing Maps, Notepad, Microsoft News \\
\midrule
AI Assistants & Microsoft Copilot, M365 Copilot \\
\bottomrule
\end{tabularx}
\end{table}

\end{document}